\documentclass[utf8]{FrontiersinHarvard} 

\usepackage{url,hyperref,lineno,microtype,subcaption}
\usepackage[onehalfspacing]{setspace}
\usepackage{multirow}
\usepackage{booktabs}
\usepackage{longtable}
\usepackage{color}

\def\keyFont{\fontsize{8}{11}\helveticabold }
\def\firstAuthorLast{Zhou {et~al.}} 
\def\Authors{Chenlin Zhou\,$^{1,\dagger}$, Han Zhang\,$^{1,2,\dagger}$, Liutao Yu\,$^{1, \dagger}$, Yumin Ye\,$^{1}$,  Zhaokun Zhou\,$^{1,3}$, Liwei Huang\,$^{1,3}$, Zhengyu Ma\,$^{1,*}$, Xiaopeng Fan\,$^{1,2}$, Huihui Zhou\,$^{1}$ and \\ 
Yonghong Tian\,$^{1,3}$}
\def\Address{$^{1}$Peng Cheng Laboratory, 
$^{2}$Harbin Institute of Technology,
$^{3}$Peking University\\
$^{\dagger}$These authors contributed equally to this work}

\def\corrAuthor{Zhengyu Ma}
\def\corrEmail{mazhy@pcl.ac.cn}

\begin{document}
\onecolumn
\firstpage{1}

\title[]{Direct Training High-Performance Deep Spiking Neural Networks: A Review of Theories and Methods} 

\author[\firstAuthorLast ]{\Authors} 
\address{} 
\correspondance{} 

\extraAuth{}%

\maketitle

\begin{abstract}

\section{}
Spiking neural networks (SNNs) offer a promising energy-efficient alternative to artificial neural networks (ANNs), in virtue of their high biological plausibility, rich spatial-temporal dynamics, and event-driven computation.
The direct training algorithms based on the surrogate gradient method provide sufficient flexibility to design novel SNN architectures and explore the spatial-temporal dynamics of SNNs. According to previous studies, the performance of models is highly dependent on their sizes. Recently, direct training deep SNNs have achieved great progress on both neuromorphic datasets and large-scale static datasets. Notably, transformer-based SNNs show comparable performance with their ANN counterparts. In this paper, we provide a new perspective to summarize the theories and methods for training deep SNNs with high performance in a systematic and comprehensive way, including theory fundamentals, spiking neuron models, advanced SNN models and residual architectures, software frameworks and neuromorphic hardware, applications, and future trends. The reviewed papers are collected in 
\href{https://github.com/zhouchenlin2096/Awesome-Spiking-Neural-Networks}{Awesome-Spiking-Neural-Networks}.

\tiny
 \keyFont{ \section{Keywords:} Deep Spiking Neural Network, Direct Training, Transformer-based SNNs, Residual Connection, Energy Efficiency, High Performance} 
\end{abstract}

\section{Introduction}

Regarded as the third generation of neural network \citep{maass1997networks}, the brain-inspired spiking neural networks (SNNs) are potential competitors to traditional artificial neural networks (ANNs) in virtue of their high biological plausibility, and low power consumption when implemented on neuromorphic hardware \citep{roy2019towards}. 
In particular, the utilization of binary spikes allows SNNs to adopt low-power accumulation (AC) instead of the traditional high-power multiply-accumulation (MAC), leading to significantly enhanced energy efficiency and making SNNs increasingly popular \citep{chen2023training}. 

There are two mainstream pathways to obtain deep SNNs: ANN-to-SNN conversion and direct training through the surrogate gradient method.
Firstly, in ANN-to-SNN conversion \citep{cao2015spiking, hunsberger2015spiking, rueckauer2017conversion, bu2021optimal, meng2022training, wang2022signed}, a pre-trained ANN is converted to an SNN by replacing the ReLU activation layers with spiking neurons and adding scaling operations like weight normalization and threshold balancing. 
This conversion process suffers from long converting time steps, which causes high computational consumption in practice. 
In addition, the converted SNNs obtained in this way are constrained by the original ANNs' architecture and are hard to adapt to dynamic signal (DVS, DAVIS, ATIS data) processing. Thus, the direct exploration of the virtues of SNNs is limited in ANN-to-SNN conversion.
Secondly, in the field of direct training, SNNs are unfolded over simulation time steps and trained with backpropagation through time \citep{lee2016training, shrestha2018slayer}. Due to the non-differentiability of spiking neurons, the surrogate gradient method is employed for backpropagation \citep{lee2020enabling, neftci2019surrogate, fang2021deep, fang2021incorporating, zhou2023spikformer}. On one hand, this direct training method can handle temporal data and also achieve decent performance on large-scale static datasets, with only a few time steps. On the other hand, it can provide sufficient flexibility for designing novel architectures specifically for SNNs and exploring the properties of SNNs directly. 
Therefore, the direct training method has received more attention recently.

Given the significant benefits and rapid advancement of directly trained deep SNNs, particularly the emergence of high-performance transformer-based SNNs, this review systematically and comprehensively summarizes the theories and methods for directly trained deep SNNs. 
Combining theory fundamentals, spiking neuron models, advanced SNN models and residual architectures, software frameworks and neuromorphic hardware, applications, and future trends, this article offers fresh perspectives into the field of SNNs. 
This review is structured as follows: Section \ref{neuron} presents the evolution and recent advancements in spiking neuron models. Section \ref{foundamentals} introduces the fundamental principles of spiking neural networks. Section \ref{Advanced Models} focuses on the most recent advanced SNN models and architectures, especially transformer-based SNNs. The performance and other details of these high-performance models are shown in table \ref{tab:performance_on_datasets}. Section \ref{frameworks} concludes the software frameworks for training SNNs and the development of neuromorphic hardware. Section \ref{sec_app} summarizes the applications of deep SNNs. Finally, Section \ref{Future trends} points out future research trends and concludes this review.

\section{Spiking Neuron Models}\label{neuron}
\begin{figure*}[t!]
	\centering
	\includegraphics[width=1.0\textwidth]{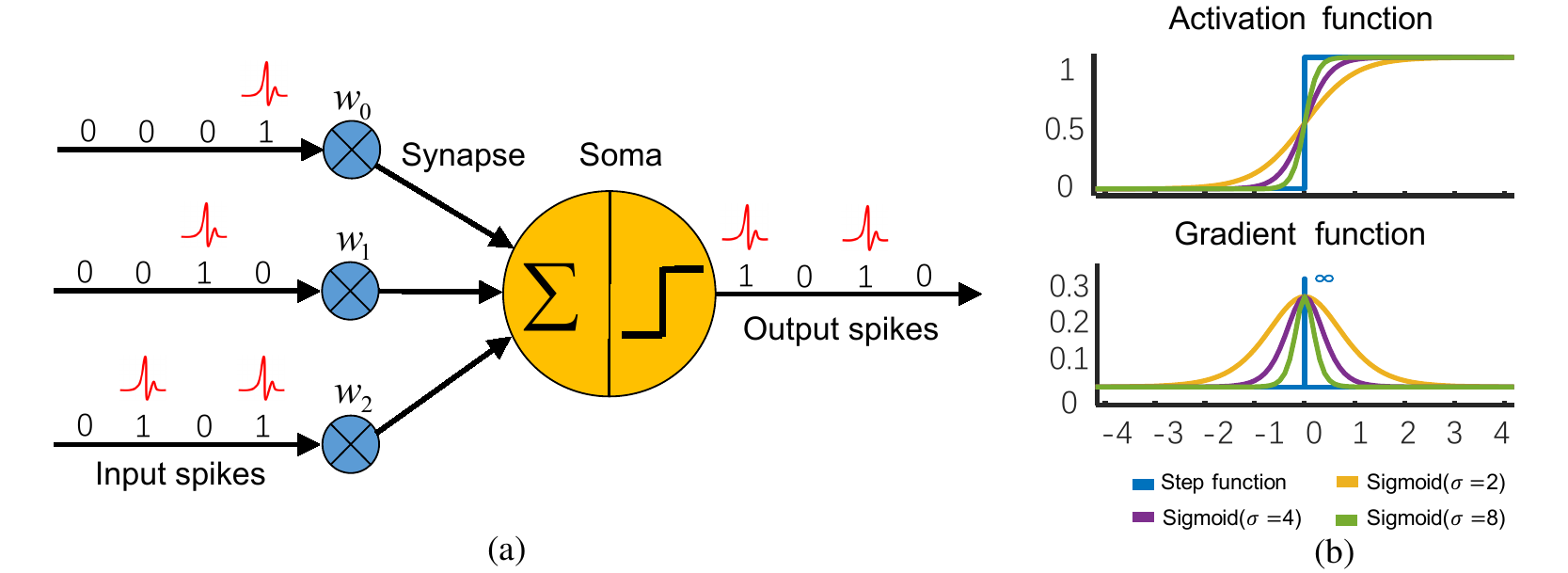}
	\caption{(a) The scheme of a spiking neuron, of which the input and output are both binary spikes. (b) The sigmoid function approximates the Heaviside activation function of a spiking neuron, and its derivative can be utilized to calculate gradients during backpropagation.}
	\label{fig:spiking_neuron}
\end{figure*}

LIF (Leaky Integrate-and-Fire) neuron is one of the most commonly used neurons in SNNs \citep{zhou2023spikformer, zhou2023spikingformer, zhou2023enhancing}, which is simple but retains biological characteristics (figure \ref{fig:spiking_neuron}a). The dynamics of LIF are described as:
\begin{equation}\label{H[t]_LIF}
    H[t]=V[t-1]+\frac{1}{\tau}\left(X[t]-\left(V[t-1]-V_{reset}\right)\right),
\end{equation}
\begin{equation}\label{S[t]_LIF}
    S[t]=\Theta\left(H[t]-V_{th}\right),
\end{equation}
\begin{equation}\label{V[t]_LIF}
    V[t]=H[t]\left(1-S[t]\right)+V_{reset}S[t],
\end{equation}
where $\tau$ in Eq. (\ref{H[t]_LIF}) is the membrane time constant, $X[t]$ is the input current at time step $t$. $V_{reset}$ represents the reset potential, $V_{th}$ represents the spike firing threshold, $H[t]$ and $V[t]$ represent the membrane potential before and after spike firing at time step $t$, respectively. $\Theta(v)$ is the Heaviside step function, if $v \geq 0$ then $\Theta(v)=1$, meaning a spike is generated; otherwise $\Theta(v)=0$. $S[t]$ represents whether a neuron fires a spike at time step $t$. 

\begin{table}[!tbp]
  \centering
  \caption{Overview of spiking neurons for direct training and their performance.}
    \begin{tabular}{p{5.5cm}<{\raggedright}p{2.8cm}<{\raggedright}p{3.0cm}<{\centering}p{1.4cm}<{\centering}p{3.0cm}<{\centering}}
    \toprule
    \multicolumn{1}{l}{Method} & \multicolumn{1}{l}{Architecture} & {Dataset}  & {Acc (\%)} & {Training} \\
    \midrule
    PLIF\citep{fang2021incorporating} & PLIF-Net &CIFAR10  &93.50 &Time dependent\\
    LTMD\citep{ltmd} & DenseNet &CIFAR10  &94.19 &Time dependent\\

    GLIF\citep{glif} & ResNet-34 &CIFAR10  &95.03 &Time dependent\\
    MLF\citep{mlf} & DS ResNet &CIFAR10  &94.25 &Time dependent\\
    LIFB\citep{lifb} & ResNet-19 &CIFAR10  &96.32 &Time dependent\\
    Deit-SNN\citep{diet-snn-tnnls} & VGG16 &CIFAR10  &93.44 &Time dependent\\
    KLIF\citep{Jiang2023KLIFAO} & CNN &CIFAR10  &92.52 &Time dependent\\
    MT-SNN\citep{mt-snn} & MT-VGG9 &CIFAR10  &94.74 &Time dependent\\
    PSN\citep{fang2023parallel} & PLIF-Net &CIFAR10  &95.32 &Parallel\\
    \midrule
    GLIF\citep{glif} & ResNet-34 &CIFAR100  &77.35 &Time dependent\\
    Deit-SNN\citep{diet-snn-tnnls} & VGG16 &CIFAR100  &69.67 &Time dependent\\
    LIFB\citep{lifb} & ResNet-19 &CIFAR100  &78.31 &Time dependent\\
    MT-SNN\citep{mt-snn} & MT-VGG9 &CIFAR100  &75.53 &Time dependent\\
    \midrule
    GLIF\citep{glif} & ResNet-34 &ImageNet  &69.09 &Time dependent\\
    Deit-SNN\citep{diet-snn-tnnls} & VGG16 &ImageNet  &69.00 &Time dependent\\
    LIFB\citep{lifb} & SEW ResNet-34 &ImageNet  &70.02 &Time dependent\\
    PSN\citep{fang2023parallel} & SEW ResNet-34 &ImageNet  &70.54 &Parallel\\
    \midrule
    PLIF\citep{fang2021incorporating} & PLIF-Net &CIFAR10-DVS  &74.80 &Time dependent\\
    GLIF\citep{glif} & ResNet-34 &CIFAR10-DVS  &78.10 &Time dependent\\
    MLF\citep{mlf} & DS ResNet &CIFAR10-DVS  &70.36 &Time dependent\\
    LTMD\citep{ltmd} & DenseNet &CIFAR10-DVS  &73.30 &Time dependent\\
    KLIF\citep{Jiang2023KLIFAO} & CNN &CIFAR10-DVS  &70.90 &Time dependent\\
    MT-SNN\citep{mt-snn} & MT-VGG9 &CIFAR10-DVS  &76.30 &Time dependent\\
    PSN\citep{fang2023parallel} & VGG &CIFAR10-DVS  &85.90 &Parallel\\
    \midrule
    PLIF\citep{fang2021incorporating} & PLIF-Net &DVS128-Gesture  &97.57 &Time dependent\\
    MLF\citep{mlf} & DS ResNet &DVS128-Gesture  &97.29 &Time dependent\\
    KLIF\citep{Jiang2023KLIFAO} & CNN &DVS128-Gesture  &94.10 &Time dependent\\
    \midrule
    \multirow{2}[0]{*}{LSNN\citep{Bellec2018LongSM}} & \multirow{2}[0]{*}{LSTM} & Sequential  & \multirow{2}[0]{*}{96.40}  & \multirow{2}[0]{*}{Time dependent} \\
     &       & MNIST &   &  \\
    ASN\citep{Yin2020EffectiveAE} & RNN &PS-MNIST  &97.90 &Time dependent\\
    SPSN\citep{spsn} & MLP &SHD  &86.89 &Parallel\\

    \bottomrule
    \end{tabular}
    \label{tab:SpikingNeurons}
\end{table}

LIF also comes with notable limitations in practical applications. For instance, LIF needs to manually adjust the hyperparameters, such as membrane time constant $\tau$ and firing threshold $V_{th}$, which constrains its expressiveness. In addition, LIF is simple in modeling, which limits the range of neuronal dynamics. Overall, there is a lack of diversity and flexibility in LIF, which calls for more advanced neuron models to enhance SNNs' performance and broaden their applications.
Table \ref{tab:SpikingNeurons} lists some recently developed spiking neuron models and their performance on typical tasks.

\subsection{Spiking Neurons with Trainable Parameters}
Based on LIF, many improved spiking neuron models with trainable parameters have been proposed, which expand the representation space of neurons through parameter learning and improve the expression ability of SNNs. Fang et al. proposed Parametric LIF (PLIF) \citep{fang2021incorporating} by using trainable membrane time constant as follows:
\begin{equation}\label{H[t]_PLIF}
    H[t]=V[t-1]+k\left(a\right)\left(X[t]-\left(V[t-1]-V_{reset}\right)\right),
\end{equation}
where $k\left(a\right)$ in Eq. (\ref{H[t]_PLIF}) denotes a clamp function and $k\left(a\right)=\frac{1}{1+exp\left(-a\right)}\in\left(0,1\right)$. The trainable membrane-related parameter of PLIF is biologically plausible, as neurons in the brain are heterogeneous. LTMD \citep{ltmd} also leverages this biological plausibility but approaches it differently by employing learnable firing thresholds. An increase in the threshold of LTMD results in a reduction of output spikes, making an SNN less sensitive to its input and thus more robust. On the contrary, a decrease in the threshold leads to an increment of output spikes, making an SNN more sensitive to its input, which is particularly beneficial for processing transient small signals. Therefore, the learnable threshold $V_{th}=\mathrm{tank}(k)$, of which $k$ is trainable, can lead to the optimal sensitivity of an SNN.

Diet-SNN\citep{diet-snn-tnnls} adopts an end-to-end gradient descent optimization algorithm to train the membrane-related parameters and firing thresholds of LIF neurons while optimizing the network weights. The trained neuron parameters selectively reduce the membrane potential, making spikes in the network sparser, thereby improving the computational efficiency of SNN. Spiking neurons with dynamic thresholds are adopted in LSNN \citep{Bellec2018LongSM}. After firing a spike each time, the firing threshold of a neuron will increase by a fixed amount, and then it will decay exponentially according to the time constant. Adaptive spiking neuron (ASN) \citep{Yin2020EffectiveAE} was proposed for sequence and streaming media tasks. In ASN, the time constant of membrane potential is trainable. In addition, similar to LSNN, the firing threshold will increase after each spike of the neuron, thus improving sparsity and efficiency. 

In KLIF\citep{Jiang2023KLIFAO}, a trainable scaling factor $k$ and a nonlinear $\mathrm{ReLU}$ activation function are inserted between charging and firing. The dynamics of KLIF can be described by Eq. (\ref{H[t]_LIF}) and Eq. (\ref{F[t]_KLIF})-(\ref{V[t]_KLIF}).
\begin{equation}\label{F[t]_KLIF}
    F[t]={\rm ReLU}\left(kH[t]\right),
\end{equation}
\begin{equation}\label{S[t]_KLIF}
    S[t]=\Theta\left(F[t]-V_{th}\right),
\end{equation}
\begin{equation}\label{V[t]_KLIF}
    V[t]=F[t]\left(1-S[t]\right)+V_{reset}S[t].
\end{equation}
Compared with LIF, KLIF can automatically adjust the membrane potential and the gradient of backpropagation within the neuron. GLIF \citep{glif} introduces a gating unit that fuses multiple biometric features, with the ratio of these features adjusted by a trainable gating factor. Moreover, inspired by various spiking patterns of brain neurons, LIFB \citep{lifb} has three modes: resting, regular spiking, and burst spiking. The density of the burst spiking can be learned automatically, which greatly enriches the representation capability of neurons. 

In addition, there are other studies trying to improve performance by multi-level firing thresholds instead of trainable parameters. To reduce the performance loss caused by the transmission of binarized spikes in the network, MT-SNN \citep{mt-snn} introduces multi-level firing thresholds. MT-SNN performs convolution operations on the binarized spikes generated by different firing thresholds and then sums them up. Similarly, MLF \citep{mlf} can also fire spikes under different firing thresholds, thus improving the performance of SNNs.

\subsection{Parallel Spiking Neurons}
A typical neuron model like LIF is time-dependent, that is, its state at time $t$ relies on its state at time $t-1$, resulting in a high computation load. Fang et al. proposed a parallel spiking neuron (PSN) to accelerate the computation by parallel computing \citep{fang2023parallel}. By eliminating the resetting process, they represent the charging process of PSN by a non-iterative equation as follows:
\begin{equation}
    H[t]=\sum_{i=0}^{T-1}W_{t,i}\cdot X[i],
\end{equation}
where $W_{t,i}$ is the weight between input $X[i]$ and membrane potential $H[t]$. For LIF neuron, $W_{t,i}=\frac{1}{\tau_{m}}(1-\frac{1}{\tau_{m}})^{t-i}$. The dynamics of PSN are as follows:
\begin{align}
    \boldsymbol{H}=\boldsymbol{W}\boldsymbol{X},\quad \boldsymbol{W}\in\mathbb{R}^{T\times T},\boldsymbol{X}\in\mathbb{R}^{T\times N} \\
    \boldsymbol{S}=\Theta(\boldsymbol{H}-\boldsymbol{B}),\quad \boldsymbol{B}\in\mathbb{R}^T,\boldsymbol{S}\in\{0,1\}^{T\times N} 
\end{align}
where $\boldsymbol{X}$ is the input, $\boldsymbol{W}$ and $\boldsymbol{B}$ are trainable weights and trainable firing thresholds, respectively. $\boldsymbol{H}$ is the membrane potential after charging, and $\boldsymbol{S}$ denotes whether a neuron spikes. $N$ and $T$ are the batch size and the number of time steps, respectively. For step-by-step serial forward computation and variable-length sequence processing, the masked PSN and the sliding PSN are also derived. 

The stochastic parallel spiking neuron (SPSN) \citep{spsn} adopts an idea similar to PSN, by removing the resetting mechanism. The neuronal dynamics of SPSN contains two parts, namely parallel leaky integrator and stochastic firing. The leaky integrator is a linear time-invariant system, which can be transformed into the Fourier domain to realize parallel computation. Stochastic firing adaptively adjusts the firing probability through trainable parameters, enhancing the network's capability to process information in a dynamic and efficient manner.

\section{Fundamentals of Spiking Neural Networks} \label{foundamentals}
\subsection{Information Coding}
To process image data through SNNs, it is essential to first encode the data into spike trains. Rate coding \citep{Impulse1926} is the most commonly used information coding method in SNNs, in which the firing rate is proportional to the intensity of the input signal and spikes are typically generated by a Poisson process \citep{Wiener2394}. To encode information more accurately, rate coding requires a longer time window, which leads to a slower information transmission rate.  In contrast, utilizing a shorter time window may result in loss of information during encoding, presenting a trade-off between speed and accuracy in information transmission.

Different from rate coding, temporal coding represents information through the timing of spikes. Time-to-first-spike (TTFS) \citep{neuralcodingzongshu, t2fsnn} stands out for its simplicity and efficiency in temporal coding, which uses the time of the first spike fired by the neuron to represent the input signal. TTFS effectively reduces the total number of spikes, thereby accelerating the computation of SNNs. TTFS algorithm can be described as follows:
\begin{equation}
    S[t]=\begin{cases}1,&\text{if}~t=\left(\frac{X_{max}-X}{X_{max}}\right)t_{max}\\0,&\text{otherwise}\end{cases} \ ,
\end{equation}
where $S[t]$ represents whether a spike is fired at time $t$ after encoding, $t_{max}$ denotes the maximum time allowed during encoding, $X$ and $X_{max}$ represent the input signal and its maximum value, respectively. In the TTFS encoding method, larger values of the input signal lead to earlier firing of spikes.

\subsection{Network Training}
\subsubsection{Surrogate Gradient}
As the core components of SNNs, neurons are essential for information processing and transmission, since spikes are fired by neurons. However, the firing of spikes involves the non-differentiable Heaviside step function, which presents a significant challenge in the direct training of SNNs. To address the non-differentiability of the Heaviside step function, Neftci et al. proposed the Surrogate Gradient (SG) \citep{neftci2019surrogate} algorithm. In SG, the Heaviside step function is adopted to generate spikes during forward propagation, and differentiable functions are adopted for gradient calculation during backpropagation. Notably, SG functions could vary according to the networks. For instance, the SG function used in SEW ResNet \citep{fang2021deep} is the derivative of the arctan function as follows:
\begin{equation}\label{sg}
    \sigma(x)=\frac{1}{\pi}\arctan(\frac{\pi}{2}\alpha x)+\frac{1}{2} \ ,
\end{equation}
\begin{equation}\label{sg_d}
    \sigma^{\prime}(x)=\frac{\alpha}{2(1+(\frac{\pi}{2}\alpha x)^2)} \ .
\end{equation}
Eq. \ref{sg_d} is the derivative of Eq. \ref{sg}. In addition, SG could be the derivative of Sigmoid (figure \ref{fig:spiking_neuron}b) \citep{zhou2023spikformer, zhou2023spikingformer, zhou2023enhancing}, tanh \citep{IMLoss}, or rectangular \citep{wu2018spatio, wu2019direct} functions, etc. To address the problem of gradient vanishing caused by surrogate gradient function with fixed parameters, Lian et al. proposed the Learnable Surrogate Gradient (LSG) \citep{LearnableSG}, in which a learnable parameter is used to adjust the gradient-available interval.

Li et al. proposed Differentiable Spike (Dspike) \citep{Dspike} as another approach to overcome the non-differentiable problem of the Heaviside function. Based on the hyperbolic tangent function, Dspike can be described as follows:
\begin{equation}
    \mathrm{Dspike}(x,b)=\frac{\tanh(b(x-0.5))+\tanh(b/2)}{2(\tanh(b/2))},\mathrm{if~}0\leq x\leq1
\end{equation}
By adjusting the parameter $b$, different backpropagation gradients can be obtained. Differentiation on Spike Representation (DSR) \citep{meng2022training} proposed by Meng et al. encodes spike trains and represents them as sub-differentiable mapping, which also avoids the non-differentiable problem during backpropagation. 

\subsubsection{Loss Function and Backpropagation}
Loss function is the key to neural network training, and different loss functions have been proposed to enhance the performance of SNNs. IM-Loss \citep{IMLoss}, for example, aims to maximize the information flow in the network. The total loss function consists of two parts, cross-entropy loss, and IM-Loss, as follows:
\begin{equation}
    \mathcal{L}_{Total}=\mathcal{L}_{CE}+\lambda\mathcal{L}_{IM},
\end{equation}
\begin{equation}
    \mathcal{L}_{IM}=\sum_{l=0}^{L}(\bar{U}_{l}-V_{th})^{2}/L,
\end{equation}
where $\bar{U}_{l}$ is the averaged membrane potential at all time steps of the $l$-th layer, and $L$ is the total number of layers. To alleviate the information loss in SNNs and reduce the quantization error, RMP-Loss \citep{guo2023rmp} is proposed to adjust the distribution of membrane potential. RecDis-SNN \citep{RecDis-SNN} adopts MDP-Loss that also adjusts the membrane potential distribution to overcome the distribution shift during network training. In addition, to improve the generalization ability of SNNs, Deng et al. proposed temporal efficient training (TET) \citep{deng2021temporal} loss function to make the network output closer to the target distribution.

\begin{figure*}[t!]
	\centering
	\includegraphics[width=1.0\textwidth]{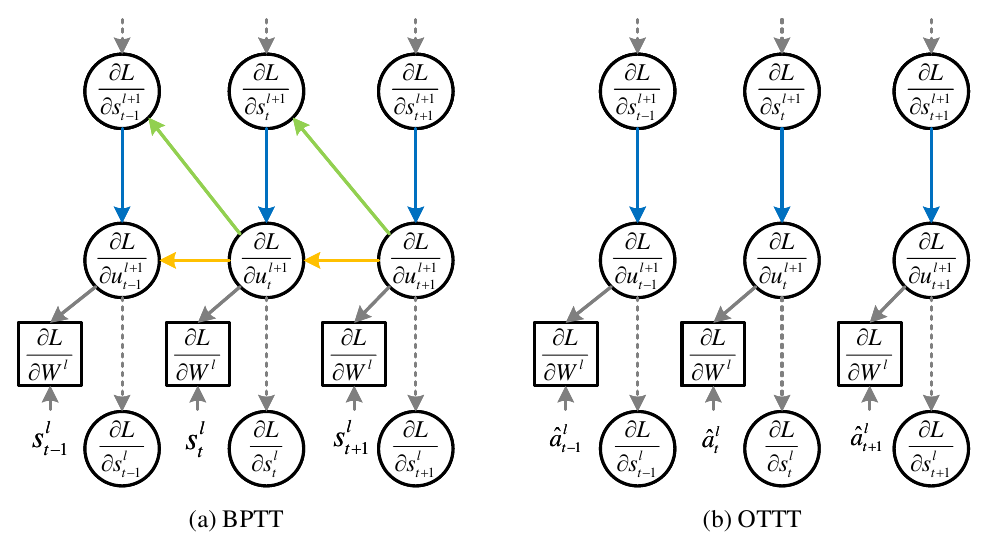}
	\caption{The backward of BPTT and OTTT.}
	\label{fig:bptt-ottt}
\end{figure*}
Distinct from ANNs, there's an additional dimension in SNNs, the temporal domain. For spiking neurons, the membrane potential in the current step depends on the membrane potential in the previous time step, that is, there is a time dependence. Thus, backpropagation in ANNs does not apply to SNNs. 
Backpropagation Through Time (BPTT) \citep{BPTT_original, BPTT}, originally developed for recurrent neural networks (RNNs), is applied to SNNs due to their similar characteristics to those of RNNs. The combination of BPTT and surrogate gradient is the basic approach in SNNs. Spatio-temporal backpropagation (STBP) \citep{wu2018spatio}, proposed by Wu et al., takes the gradient update in both the spatial domain and temporal domain into account to train SNNs. However, the additional time dimension exposes BPTT and STBP to the problem of requiring a large amount of training memory and training time. Therefore, Xiao et al. proposed an online training through time (OTTT) \citep{OTTT} algorithm derived from BPTT, which only requires constant training memory consumption agnostic to time steps, and reduces the significant memory costs compared to BPTT. The backward of BPTT and OTTT are shown in figure \ref{fig:bptt-ottt}. Another efficient backpropagation method, Spatial Learning Through Time (SLTT) \citep{SLTT}, ignores the unimportant routes in the computational graph during backpropagation, to reduce training memory consumption and training time. However, although OTTT and SLTT show better training memory consumption than BPTT, direct training high-performance SNNs are still dominated by the combination of BPTT and surrogate gradient, such as SGLFormer \citep{zhang2024sglformer}, Spikformer \citep{zhou2023spikformer}, etc. Thus, it's essential to investigate direct training methods offering both high effectiveness and efficiency.

\subsubsection{Batch Normalization}
In SNNs, batch normalization is an indispensable component, especially in the context that deep SNNs are difficult to train and converge, compared to ANNs. To mitigate the degradation problems of SNNs, Zheng et al. proposed threshold-dependent batch normalization (tdBN) \citep{zheng2021going}, which is described as follows:
\begin{equation}
    \hat{X}_{k}=\gamma_{k}\frac{\alpha V_{th}(X_{k}-\mu)}{\sqrt{\sigma^2+\epsilon}}+\beta_{k},
\end{equation}
where $\alpha$ is a hyperparameter, $V_{th}$ is the firing threshold of the neuron, $X_k$ is the feature of the $k$-th channel, $\gamma_k$ and $\beta_k$ are trainable parameters, $\mu$ and $\sigma^2$ are mean and variance respectively, $\epsilon$ is a tiny constant. Temporal effective batch normalization (TEBN) \citep{TEBN} regularizes the temporal distribution, by adopting batch normalization with different parameters at different time steps.
Batch normalization through time (BNTT) \citep{BNTT} proposed by Kim et al. is similar to TEBN, which also adopts different batch normalization parameters for feature maps at different time steps. Moreover, Guo et al. applied batch normalization inside the LIF neuron to normalize the distribution of membrane potentials before firing spikes \citep{guo2023membrane}.

\section{SNN Architecture developments} \label{Advanced Models}
This review focuses on the most recent SNN models. 
Recently, the evolution of residual blocks enhances both the size and performance of deep SNNs significantly. In addition, combining SNNs with transformer architecture has broken the bottleneck of SNNs' performance. Therefore, this review focuses on the application of two kinds of architectures in direct training deep SNNs: transformer structures \ref{sec_Trans} and the residual connections \ref{sec_RL}. 
Table \ref{tab:performance_on_datasets} summarizes their performance on mainstream datasets (ImageNet-1K, CIFAR10, CIFAR100, DVS128 Gesture, CIFAR10-DVS).

\begin{figure*}[!t]
	\centering
	\includegraphics[width=1.0\textwidth]{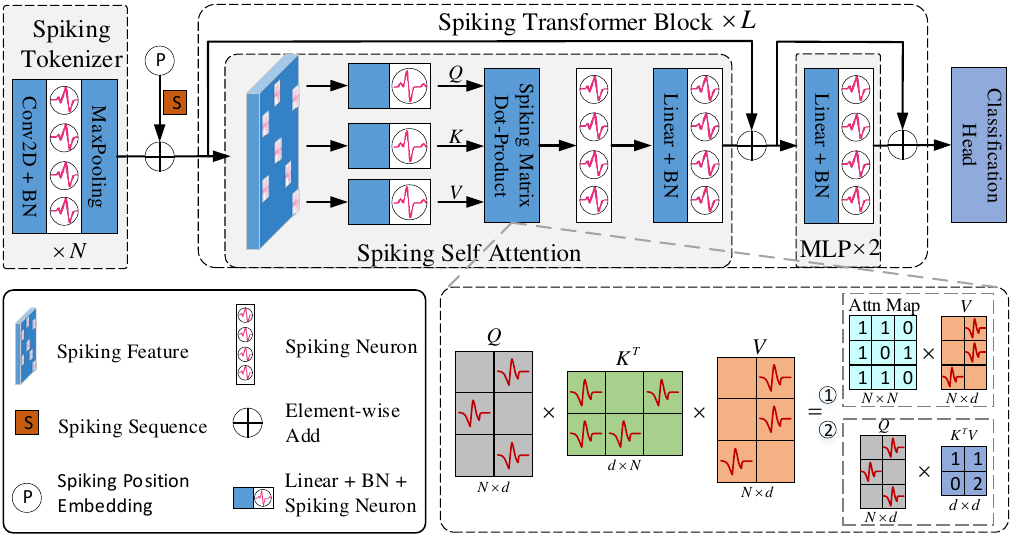}
	\caption{The overview of Spiking Transformer (Spikformer).}
	\label{fig:spikformer}
\end{figure*}

\subsection{Transformer-based Spiking Neural Networks}\label{sec_Trans}
Transformer, originally designed for natural language processing \citep{vaswani2017attention}, has achieved great success in many computer vision tasks, including image classification \citep{dosovitskiy2020image, yuan2021tokens}, object detection \citep{carion2020end, zhu2020deformable, liu2021swin} and semantic segmentation \citep{wang2021pyramid, yuan2022volo}. While convolution-based models mainly rely on inductive bias and focus on adjacent pixels, transformer structures use self-attention to capture the relation among spiking features globally, which enhances the performance effectively.

\begin{table*}[htbp]
    \centering
    \caption{Overview of direct training deep SNNs and their performance on ImageNet, CIFAR10, CIFAR100, DVS128-Gesture, CIFAR10-DVS.}
    \label{tab:performance_on_datasets}
    \begin{tabular}{llp{0.8cm}<{\centering}p{0.7cm}<{\centering}cp{1.4cm}<{\centering}}
    \toprule
    Method &{Architecture} & {Param (M)} & Time steps  & {Dataset} & {Top-1 Acc ($\%$)} \\ 
    \midrule

    
    
                                     
    {\multirow{1}{*}{Spiking ResNet\citep{hu2021spiking}}} 
                                     &ResNet-50 &25.56 & 350 &ImageNet & 72.75 \\
    \multirow{1}{*}{SEW ResNet\citep{fang2021deep}}  
                                                     & SEW-ResNet-152 &60.19   &4 &ImageNet & 69.26 \\

    {\multirow{1}{*}{{MS-ResNet\citep{hu2021advancing}}}}
                                   &{MS-ResNet-104}&77.28 & 5 &ImageNet & {76.02}\\

    {\multirow{1}{*}{{Att MS-ResNet\citep{10032591}}}}

                                   &{Att-MS-ResNet-104}&78.37 & 4 &ImageNet & {77.08}\\

    {\multirow{1}{*}{{Spikformer\citep{zhou2023spikformer}}}}

                                   &{Spikformer-8-768}&66.34 & 4 &ImageNet & {74.81}\\

    {\multirow{1}{*}{Spikingformer\citep{zhou2023spikingformer}}}

                                   &{Spikingformer-8-768}&66.34 & 4 &ImageNet & {75.85}\\

    {\multirow{1}{*}{{CML\citep{zhou2023enhancing}}}}
                                   &{Spikformer-8-768}&66.34 & 4 &ImageNet & {77.34}\\


    Spike-driven Transformer & \multirow{2}[0]{*}{S-Transformer-8-768} & \multirow{2}[0]{*}{66.34} & \multirow{2}[0]{*}{4} & \multirow{2}[0]{*}{ImageNet}  & \multirow{2}[0]{*}{77.07} \\
    \citep{yao2023spikedriven} &       &  &   &   &  \\

    {\multirow{1}{*}{{SpikingResformer\citep{shi2024spikingresformer}}}}
                                   &{SpikingResformer-L}&60.38 & 4 &ImageNet & {79.40}\\

    Spike-driven Transformer V2 & \multirow{2}[0]{*}{Meta-SpikeFormer} & \multirow{2}[0]{*}{55.40} & \multirow{2}[0]{*}{4} & \multirow{2}[0]{*}{ImageNet}  & \multirow{2}[0]{*}{80.00} \\
    \citep{yao2024spikedriven} &       &  &   &   &  \\

    {\multirow{1}{*}{{Spikformer V2\citep{zhou2024spikformerv2}}}}
                                   &{Spikformer V2-8-512}&51.55 & 4 &ImageNet & {80.38}\\
    
    {\multirow{1}{*}{{SGLFormer\citep{zhang2024sglformer}}}}
                                   &{SGLFormer-8-768}&64.02 & 4 &ImageNet & {83.73}\\
    {\multirow{1}{*}{{QKFormer\citep{zhou2024qkformer}}}}
                                   &{HST-10-768}&64.96 & 4 &ImageNet & {85.65}\\

    \midrule
    Hybrid training\citep{rathi2019enabling} & VGG-11 &9.27  &125 &CIFAR10 &92.22 \\
    STBP-tdBN\citep{zheng2021going} &ResNet-19 &12.63 & 4 &CIFAR10 & 92.92\\
    TET\citep{deng2021temporal}  &ResNet-19 &12.63 & 4 &CIFAR10 & {94.44}\\
    MS-ResNet\citep{hu2021advancing} &MS-ResNet-110 &- & 4 &CIFAR10 & {92.12}\\
    {\multirow{1}{*}{{Spikformer\citep{zhou2023spikformer}}}}
                        & {Spikformer-4-384} &9.32 & 4 &CIFAR10 & {95.51}\\
                        
    {\multirow{1}{*}{{Spikingformer\citep{zhou2023spikingformer}}}}
                        & {Spikingformer-4-384} &9.32 & 4 &CIFAR10 & {95.81}\\

    {\multirow{1}{*}{{CML\citep{zhou2023enhancing}}}}
                        & {Spikformer-4-384} &9.32 & 4 &CIFAR10 & {96.04}\\

    
    Spike-driven Transformer & \multirow{2}[0]{*}{S-Transformer-2-512} & \multirow{2}[0]{*}{10.23} & \multirow{2}[0]{*}{4} & \multirow{2}[0]{*}{CIFAR10}  & \multirow{2}[0]{*}{95.60} \\
    \citep{yao2023spikedriven} &       &  &   &   &  \\
    
    {\multirow{1}{*}{{SGLFormer\citep{zhang2024sglformer}}}}
                                   &{SGLFormer-4-384}&8.85 & 4 &CIFAR10 & {96.76}\\

    \midrule
    Hybrid training\citep{rathi2019enabling} & VGG-11 &9.27  &125 &CIFAR100 &67.87 \\
    STBP-tdBN\citep{zheng2021going} &ResNet-19 &12.63 & 4 &CIFAR100 & 70.86\\
    TET\citep{deng2021temporal}  &ResNet-19 &12.63 & 4 &CIFAR100 & {74.47}\\
    
    {\multirow{1}{*}{{Spikformer\citep{zhou2023spikformer}}}}
                        & {Spikformer-4-384} &9.32 & 4 &CIFAR100 & {78.21}\\

    {\multirow{1}{*}{{Spikingformer\citep{zhou2023spikingformer}}}}
                        & {Spikingformer-4-384} &9.32 & 4 &CIFAR100 & {79.21}\\

    {\multirow{1}{*}{{CML\citep{zhou2023enhancing}}}}
                        & {Spikformer-4-384} &9.32 & 4 &CIFAR100 & {80.02}\\

    Spike-driven Transformer & \multirow{2}[0]{*}{S-Transformer-2-512} & \multirow{2}[0]{*}{10.28} & \multirow{2}[0]{*}{4} & \multirow{2}[0]{*}{CIFAR100}  & \multirow{2}[0]{*}{78.4} \\
    \citep{yao2023spikedriven} &       &  &   &   &  \\

        
    {\multirow{1}{*}{{SGLFormer\citep{zhang2024sglformer}}}}
                                   &{SGLFormer-4-384}&8.88 & 4 &CIFAR100 & {82.26}\\

    \midrule
    SEW-ResNet\citep{hu2021advancing} &SEW-ResNet &- & 16 &DVS128-Gesture & {97.9}\\
    tdBN \citep{zheng2021going} &ResNet &- & 40 &DVS128-Gesture & {96.9}\\

    {\multirow{1}{*}{{Spikformer\citep{zhou2023spikformer}}}}
                        & {Spikformer-2-256} &2.57 & 16 &DVS128-Gesture & {98.3}\\
                        
    {\multirow{1}{*}{{Spikingformer\citep{zhou2023spikingformer}}}}
                        & {Spikingformer-2-256} &2.57 & 16 &DVS128-Gesture & {98.3}\\

    {\multirow{1}{*}{{CML\citep{zhou2023enhancing}}}}
                        & {Spikformer-2-256} &2.57 & 16 &DVS128-Gesture & {98.6}\\


    Spike-driven Transformer & \multirow{2}[0]{*}{S-Transformer-2-256} & \multirow{2}[0]{*}{2.57} & \multirow{2}[0]{*}{16} & \multirow{2}[0]{*}{DVS128-Gesture}  & \multirow{2}[0]{*}{99.3} \\
    \citep{yao2023spikedriven} &       &  &   &   &  \\

    {\multirow{1}{*}{{STSA\citep{ijcai2023p344}}}}
                        &STSFormer-2-256 &1.99 &16 &DVS128-Gesture & 98.72\\
                                
    {\multirow{1}{*}{{SGLFormer\citep{zhang2024sglformer}}}}
                                   &{SGLFormer-3-256}&2.17 & 16 &DVS128-Gesture & {98.6}\\

    \midrule
    SEW-ResNet\citep{hu2021advancing} &SEW-ResNet &- & 16 &CIFAR10-DVS & {74.4}\\

    {\multirow{1}{*}{{Spikformer\citep{zhou2023spikformer}}}}
                        & {Spikformer-2-256} &2.57 & 16 &CIFAR10-DVS & {80.9}\\

    {\multirow{1}{*}{{Spikingformer\citep{zhou2023spikingformer}}}}
                        & {Spikingformer-2-256} &2.57 & 16 &CIFAR10-DVS & {81.3}\\

    {\multirow{1}{*}{{CML\citep{zhou2023enhancing}}}}
                        & {Spikformer-2-256} &2.57 & 16 &CIFAR10-DVS & {80.9}\\


    Spike-driven Transformer & \multirow{2}[0]{*}{S-Transformer-2-256} & \multirow{2}[0]{*}{2.57} & \multirow{2}[0]{*}{16} & \multirow{2}[0]{*}{CIFAR10-DVS}  & \multirow{2}[0]{*}{80.0} \\
    \citep{yao2023spikedriven} &       &  &   &   &  \\

    {\multirow{1}{*}{{STSA\citep{ijcai2023p344}}}}
                        &STSFormer-2-256 &1.99 &16 &CIFAR10-DVS & 79.93\\

    {\multirow{1}{*}{{SGLFormer\citep{zhang2024sglformer}}}}
                               &{SGLFormer-3-256}&2.58 & 10 &CIFAR10-DVS & {82.9}\\
    \bottomrule
    \end{tabular}%
\end{table*}

To adopt transformer structure in SNNs, Zhou et al. designed a novel spike-form self-attention named Spiking Self Attention (SSA) \citep{zhou2023spikformer}, using sparse spike-form Query, Key and Value without softmax operation. The calculation process of SSA is formulated as follows:
\begin{equation}
Q=\mathrm{S N}_Q\left(\mathrm{BN}\left(X W_Q\right)\right), K=\mathrm{S N}_K\left(\mathrm{BN}\left(X W_K\right)\right), V=\mathrm{S N}_V\left(\mathrm{BN}\left(X W_V\right)\right),
\end{equation}
\begin{equation}
\operatorname{SSA}^{\prime}(Q, K, V)=\mathrm{S N}\left(Q K^{\mathrm{T}} V * s\right),
\end{equation}
\begin{equation}
\operatorname{SSA}(Q, K, V)=\mathrm{S N}\left(\operatorname{BN}\left(\operatorname{Linear}\left(\operatorname{SSA}^{\prime}(Q, K, V)\right)\right)\right),
\end{equation}
where $Q, K, V \in \mathbb{R}^{T \times N \times D}$. The spike-form Query ($Q$), Key ($K$), and Value ($V$) are computed by learnable layers. $s$ is a scaling factor, which can be fused into the next spiking neuron in practice. Therefore, the calculation of SSA avoids multiplication, meeting the property of SNNs.
Based on the SSA, Zhou et al. developed a spiking transformer named Spikformer \citep{zhou2023spikformer}, which is shown in figure \ref{fig:spikformer}. As the first transformer-based SNN model, Spikformer achieves 74$\%$ accuracy on ImageNet-1k, showing great performance potential.

\cite{zhou2023spikingformer} discussed the non-spike computation problem (integer-float multiplications) of Spikformer \citep{zhou2023spikformer} and SEW-ResNet \citep{fang2021deep}, which is caused by Activation-after-addition shortcut. Spikingformer \citep{zhou2023spikingformer} was proposed with the Pre-activation shortcut to avoid the non-spike computation problem in synaptic computing. Experimental Analysis has shown that Spikingformer has only about 43$\%$ energy consumption compared with Spikformer in synaptic computing, with only accumulation operations and lower fire rates. 
CML \citep{zhou2023enhancing} designed a downsampling structure specifically for SNNs to solve the imprecise gradient backpropagation problem of most state-of-the-art deep SNNs (including Spikformer). CML achieved 77.34$\%$ on ImageNet, significantly enhancing the performance of transformer-based SNNs. 
All the architectures above are based on SSA with computational complexity of $O(N^2 d)$ or $O(Nd^2)$, 
while \cite{yao2023spikedriven} designed a novel Spike-Driven Self-Attention (SDSA) with linear complexity regarding both the number of tokens and channels. SDSA uses only mask and addition operations without any multiplication, thus having up to 87.2$\times$ lower computation energy than the vanilla SSA. In addition, the Spike-driven Transformer based on SDSA has achieved 77.1$\%$ accuracy on ImageNet-1k. \cite{ijcai2023p344} proposed an SNN-based spatial-temporal self-attention (STSA) mechanism, which could calculate the feature dependence across the time and space domains.
{
\cite{shi2024spikingresformer} proposed Dual Spike Self-Attention (DSSA) with a reasonable scaling method, achieving 79.40$\%$ top-1 accuracy on ImageNet-1K.
\cite{yao2024spikedriven} proposed Spike-driven Transformer v2 which explored the impact of structure, spike-driven self-attention, and skip connection on its performance to inspire the next-generation transformer-based neuromorphic chip designs.
\cite{zhou2024spikformerv2} developed a Spiking Convolutional Stem (SCS) with supplementary layers to enhance the architecture of Spikformer, achieving 80.38$\%$ accuracy on ImageNet-1k. 
\cite{zhang2024sglformer} proposed a Spiking Global-Local-Fusion Transformer (SGLFormer), which enables efficient information processing on both global and local scales, by integrating transformer and convolution structures in SNNs. SGLFormer achieved a groundbreaking top-1 accuracy of 83.73$\%$ on ImageNet-1k with 64M parameters. 
\cite{zhou2024qkformer} proposed QKFormer, a novel hierarchical spiking transformer using Q-K attention, which can easily model the importance of token or channel dimensions with binary values and has linear complexity to \#tokens (or \#channels).
QKFormer achieved a significant milestone, surpassing 85\% top-1 accuracy on ImageNet with 4 time steps using the direct training approach.
}

{
Biological realistic models tend to model neural networks with high biological plausibility to simulate the complex biological mechanism of the brain. 
It often lacks the consideration of computational efficiency and performance optimization on general application tasks.
Traditional ANNs often prioritize task performance over biological realism and computational energy consumption. 
SNNs have great potential to own the characteristics of biological plausibility, low computational energy consumption, and high task performance simultaneously. Especially, several direct training Transformer-based SNNs have broken through 80\% top-1 accuracy on ImageNet-1K, which instills great optimism in the application of SNNs.
}

\subsection{Residual Architectures in Spiking Neural Networks}\label{sec_RL}
\begin{figure*}[!t]
	\centering
	\includegraphics[width=1.0\textwidth]{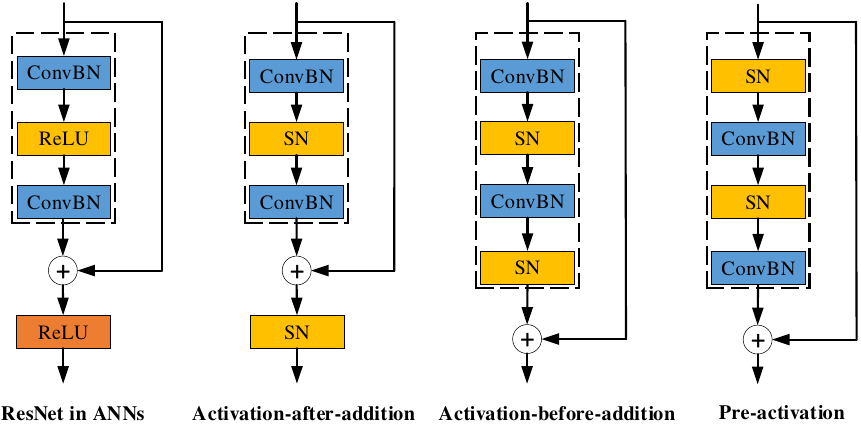}
	\caption{The overview of residual learning architectures.}
	\label{fig: residual}
\end{figure*}
Residual block is the fundamental block in both deep ANNs and SNNs. As shown in figure \ref{fig: residual}, there are mainly three residual shortcut types in SNNs: Activation-after-addition, Activation-before-addition, and Pre-activation. Both advantages and disadvantages of these three types are concluded in table \ref{tab: residual}. \textbf{Activation-after-addition shortcut} simply replaces ReLU activation layers in the standard residual block with spiking neurons, such as Spiking ResNet \citep{hu2021spiking} and MPBN\citep{guo2023membrane}. SNNs with this simple design suffer from performance degradation and gradient vanishing/exploding. For example, the deeper
34-layer Spiking ResNet has lower test accuracy than the shallower 18-layer Spiking ResNet. As the layer increases, the test accuracy of Spiking ResNet decreases \citep{fang2021deep}.
To solve the degradation problem in the Activation-after-addition shortcut, \textbf{Activation-before-addition shortcut} is proposed in SEW-ResNet \citep{fang2021deep}, which extended directly trained SNNs to 100 layers for the first time. This structure has been widely used, such as in Spikformer\citep{zhou2023spikformer}, PLIF\citep{fang2021incorporating}, PSN \citep{fang2023parallel}. This design mitigates the vanishing/exploding gradient problem and could train deeper SNN. However, the blocks in this shortcut will result in positive integers, which leads to non-spike computations (integer-float multiplications) in synaptic computing (like convolutional layer, linear layer) \citep{zhou2023spikingformer}. \textbf{Pre-activation shortcut} could be traced back to the $Activation\text{-}Conv\text{-}Bn$ paradigm, which is a fundamental building block in Binary Neural Networks (BNNs) \cite{liu2018bi,guo2021boolnet,zhang2022pokebnn,liu2020reactnet}. Some representative SNNs that use the Pre-activation shortcut include MS-ResNet\citep{hu2021advancing}, Spikingformer\citep{zhou2023spikingformer}, Spike-driven transformer\citep{yao2023spikedriven}. MS-ResNet directly trained convolution-based SNNs to successfully extend the depth up to 482 layers on CIFAR10 without experiencing degradation problems, effectively verifying the feasibility of this way. Spikingformer\citep{zhou2023spikingformer} showed that the Pre-activation shortcut can effectively avoid non-spike computations, and thus has lower energy consumption than the previous shortcut in synaptic computing, through avoiding integer-float multiplication problems and with a lower firing rate. However, the Pre-activation shortcut requires dense transmission of floats in the residual branch. 

\newcolumntype{L}[1]{>{\raggedright\let\newline\\\arraybackslash\hspace{0pt}}m{#1}}
\newcolumntype{C}[1]{>{\centering\let\newline\\\arraybackslash\hspace{0pt}}m{#1}}
\newcolumntype{R}[1]{>{\raggedleft\let\newline\\\arraybackslash\hspace{0pt}}m{#1}}
\begin{table}[!b]
  \vspace{-0.5cm}
  \centering
  \caption{Features of various residual learning architectures.}
    \begin{tabular}{L{3.5cm}L{4.5cm}L{4.2cm}L{3.5cm}}
    \toprule
     \textbf{Features} & \textbf{Activation-after-addition} & \textbf{Activation-before-addition} & \textbf{Pre-activation}  \\
   \midrule
    Element addition  & Spike added to floats & Integer added to spike  & Floats added to floats \\
    \midrule
    Gradient vanishing/ exploding  & Yes  & No  & No \\
    \midrule
    Synaptic computing type & Spike computing  & Multiplication between \quad \quad \quad sparse integer and floats & Spike computing \\
    \midrule
    Data transmission in residual branch & Spike  & Sparse integer  & Floats \\
    \midrule
    \end{tabular}%
    \label{tab: residual}
\end{table}%

Overall, the residual learning suitable for the properties of SNNs needs further exploration. In our opinion, the Activation-after-addition shortcut with gradient problem is not suitable for directly training deep SNNs, but is feasible in the field of ANN-to-SNN conversion (ANN2SNN). Activation-before-addition shortcut has some alternatives to ensure the properties of SNNs by slightly sacrificing the performance, such as using AND or IAND to replace ADD in the aggregation operation. Pre-activation shortcut needs further analyses of the effects of float transmission, and more efforts to exploit its advantages through collaborative hardware optimization and design.

\subsection{Others}
{
Besides the above-mentioned architectures, some other interesting research topics are also worthy of attention, such as Spiking RNN/LSTM, LSM, etc.
Deep Liquid State Machine (LSM) \citep{wang2016d} explored the power of recurrent spiking networks and deep architectures.
\cite{soures2019deep} proposed a novel deep LSM to capture dynamic information over multiple time-scales with a combination of randomly connected layers and unsupervised layers.
\cite{hamilton2019spike} demonstrated the nonlinear dynamics of spiking neurons can be used to implement low-level graph operations.
\cite{zhu2022spiking} proposed end-to-end Spiking Graph Convolutional Networks (GCNs) that integrate the embedding of GCNs with the biofidelity characteristics of SNNs.
\cite{bellec2020solution} and \cite{bohnstingl2022biologically} explored the architectures and online-training methods of recurrent spiking neural networks.
\cite{ren2023spikepoint} proposed a novel end-to-end point-based SNN architecture, which excels at processing sparse event cloud data, effectively extracting both global and local features through a singular-stage structure.
}

\section{Software Frameworks and Neuromorphic Hardware for Spiking Neural Networks} 
\label{frameworks}
\subsection{Software Frameworks for Training Spiking Neural Networks}\label{software}
Software frameworks play a crucial role in propelling the advancement of deep learning. Deep learning frameworks such as PyTorch \citep{paszke2019pytorch} and TensorFlow \citep{199317} leverage low-level languages like C++ libraries for high-performance acceleration on the backend, while offering user-friendly front-end application programming interfaces (APIs) implemented in high-level languages like Python. These frameworks significantly ease the workload of constructing and training ANNs, making substantial contributions to the growth of deep learning research. However, these deep learning frameworks are primarily designed for ANNs. With the development of large-scale brain-inspired neural networks, many related frameworks have emerged, facilitating the modeling and efficient computation of large-scale SNNs.

One category of frameworks includes brain simulators such as NEURON \citep{hines1997neuron} and Brian \citep{goodman2009brian}, which not only enhance the scalability and computational efficiency of models but also encompass cognitive functions such as perception, decision-making, and reasoning. The SNNs constructed by these frameworks exhibit a high degree of biological plausibility, making them suitable for studying the functionalities of real neural systems. They support biologically interpretable learning rules such as Spike-Timing-Dependent Plasticity (STDP) \citep{bi1998synaptic}, playing a significant role in advancing the field of neuroscience. However, these frameworks lack core computational functionalities required for deep learning, such as automatic differentiation, rendering them incapable of performing machine learning tasks. 
Another category of brain-inspired computing frameworks comprises deep spiking computation frameworks. Deep SNNs involve a substantial amount of matrix operations across spatial and temporal dimensions, a variety of neurons, neuromorphic datasets, and deployments on neuromorphic chips. The modeling and application processes are complex, and achieving high-performance acceleration is challenging. To address these issues, spiking deep learning frameworks need to support the construction, training, and deployment of deep SNNs, and be capable of acceleration based on spike operations. Frameworks such as BindsNET \citep{hazan2018bindsnet}, NengoDL \citep{rasmussen2019nengodl}, SpykeTorch \citep{mozafari2019spyketorch}, \href{https://doi.org/10.5281/zenodo.4422025}{Norse}, \href{https://github.com/fzenke/spytorch}{SpyTorch}, SNNTorch \citep{eshraghian2023training} and SpikingJelly \citep{fang2023spikingjelly} have been developed. They utilize simple spiking neurons to reduce computational complexity, making them suitable for machine learning research. Among them, BindsNet \citep{hazan2018bindsnet} primarily focuses on machine learning and reinforcement learning; NengoDL \citep{rasmussen2019nengodl} converts ANNs to obtain deep SNNs but does not support direct training of SNNs using surrogate gradient methods; SpyTorch is a demonstrative framework that only provides basic surrogate gradient examples; SpyTorch \citep{mozafari2019spyketorch} introduces a new type of surrogate gradient method named SuperSpike. These frameworks can implement some simple machine learning and reinforcement learning models, but they still lack deep learning capabilities for SNNs. Norse is attempting to introduce the sparse and event-driven characteristics of SNNs and supports many typical spiking neuron models. It is in the development stage and has not been officially released yet. SNNTorch supports some variants of online backpropagation algorithms that are more biologically plausible and support large-scale SNN computation. 
SpikingJelly \citep{fang2023spikingjelly} is a full-stack toolkit for preprocessing neuromorphic datasets, building deep SNNs, optimizing their parameters, and deploying SNNs on neuromorphic chips, which shows remarkable extensibility and flexibility, enabling users to accelerate custom models at low costs through multilevel inheritance and semiautomatic code generation. In summary, the development of existing software frameworks is essentially in its early stages, and there is still a long way to go in terms of functionality enhancement and performance optimization.
\begin{table*}[!tbp]
  \centering
  \caption{Overview of typical neuromorphic hardware. (SNN TA: SNN Training Accelerator, Arch: Architecture)}
  \label{tab:neuromorphic_hardware}
        \begin{tabular}{llllp{1.5cm}<{\centering}p{1.1cm}<{\centering}}
    \toprule
     Chip & Developer & Network & Function & Arch & Scale \\
   \midrule
   \multicolumn{6}{c}{Hybrid Digital-Analog} \\
   \midrule
   BrainScaleS \citep{schemmel2010wafer} & Heidelberg Uni & SNNs & Training & NMA & Large\\
   Neurogrid \citep{benjamin2014neurogrid} & Stanford & SNNs & Inference & NMA & Large\\
   ROLLS \citep{qiao2015reconfigurable} & UZH & SNNs & Training & NMA & Small\\
    DYNAPs \citep{moradi2017scalable} & UZH & SNNs & Inference & NMA & Small\\ 
   Memristor-based \citep{zhang2021hybrid} & - & ANNs\slash SNNs & Inference & CIM & Small\\
   BrainScaleS-2 \citep{pehle2022brainscales} & Heidelberg Uni & ANNs\slash SNNs & Training & NMA & Large\\
   
   \midrule
   \multicolumn{6}{c}{Digital} \\
   \midrule
   SpiNNaker \citep{painkras2013spinnaker} & UoM & SNNs & Training & NMA & Large\\ 
   SpiNNaker 2 & UoM & ANNs\slash SNNs & Training & NMA & Large\\
   TrueNorth \citep{akopyan2015truenorth} & IBM & SNNs & Inference & NMA & Large\\ 
   Darwin \citep{shen2016darwin} & ZJU & SNNs & Inference & NMA & Small\\
   Darwin II \citep{ma2017darwin} & ZJU & SNNs & Training & NMA & Large\\
   DeepSouth \citep{wang2017neuromorphic} & Westwell & SNNs & Inference & NMA & Large\\ 
   
   Intel SNN chip \citep{chen20184096} & Intel & SNNs & Training & NMA & Large\\ 
   ODIN \citep{frenkel20180} & K.U.Leuven & SNNs & Training & NMA & Small\\
   Loihi \citep{davies2018loihi} & Intel & SNNs & Training & NMA & Large\\ 
   Tianjic \citep{pei2019towards} & Tsinghua & ANNs\slash SNNs & Training & NMA & Large\\
   MorphIC \citep{frenkel2019morphic} & UZH & SNNs & Training & NMA & Small\\
   Flash-based \citep{wu2020compact} & - & ANNs\slash SNNs & Inference & CIM & Small\\
   Loihi II \citep{davies2021taking} & Intel & SNNs & Training & NMA & Large\\
   Y. Kuang et al. \citep{kuang202164k} & PKU &  ANNs\slash SNNs & Inference & NMA & Large\\
   H2Learn \citep{liang2021h2learn} & UCSB & SNNs & Training & SNN TA & Large\\
   SATA \citep{yin2022sata} & Yale & SNNs & Training & SNN TA & Small\\
    \midrule
    \end{tabular}%
\end{table*}%


\subsection{Neuromorphic Hardware for Spiking Neural Networks} \label{Hardware}
Neuromorphic hardware provides computational power for neural network models, playing a crucial role in large-scale brain-like neural networks. Efficient hardware can significantly accelerate the training, evaluation, iteration, and real-world applications of large-scale brain-like models. In comparison to general-purpose processors, deep learning chips and brain-like chips are specialized chips that focus on the computational efficiency of deep learning tasks and brain-like computing tasks, aiming to achieve better power/performance/area ratios. From an architectural perspective, current brain-like chips can be mainly divided into two categories: analog-digital hybrid circuits and fully digital circuits (table \ref{tab:neuromorphic_hardware}). Current deep learning chips, like general CPUs, are based on the Von Neumann architecture, with separate computing and storage units. Brain-like chips enhance computational efficiency by designing efficient storage and computation hierarchy, enabling parallel data flow and efficient reuse, thus improving computational efficiency.

Inspired by the simultaneous computation and storage capabilities of the brain's neural system, brain-inspired chips often adopt near-memory(NMA) or compute-in-memory architectures(CIM), incorporating closely coupled computational and storage resources within each computing core \citep{akopyan2015truenorth, pei2019towards}. Efficient intra-chip and inter-chip interconnects enable large-scale computational parallelism  and high local memory, reducing computational power consumption. 
The near-memory computing architecture refers to the separation of memory storage and computation in each processing unit, but with proximity. Key chips in this category include IBM’s TrueNorth \citep{akopyan2015truenorth}, Intel’s Loihi \citep{davies2018loihi, davies2021taking}, the University of Manchester’s SpiNNaker \citep{painkras2013spinnaker}, Stanford University’s Neurogrid \citep{benjamin2014neurogrid}, Heidelberg University’s BrainScaleS \citep{schemmel2010wafer, pehle2022brainscales}, Tsinghua University’s Tianji Chip \citep{pei2019towards}, and Zhejiang University’s Darwin Chip \citep{shen2016darwin, ma2017darwin}. They utilize characteristics of brain-like spiking computation such as sparsity, spike summation, and asynchronous event-driven processing to achieve ultra-low power consumption, currently mainly supporting model inference and local online learning based on STDP, Hebb, etc. For instance, ROLLS \citep{qiao2015reconfigurable}, ODIN \citep{frenkel20180}, and MorphIC \citep{frenkel2019morphic} support spike-driven synaptic plasticity (SDSP) rules, and Loihi adds a learning module for STDP rules. In SpiNNaker \citep{painkras2013spinnaker} and BrainScaleS \citep{schemmel2010wafer}, STDP learning is exhibited through timestamp recording and learning circuits. In their next generations \citep{pehle2022brainscales}, more flexible learning rules are possible due to the presence of embedded programmable units. Tsinghua University's Tianji Chip, as the first chip to support the fusion of SNN and ANN computation, improves accuracy based on ANN, and achieves rich dynamics, high efficiency, and robustness based on SNN. This mode is also adopted by BrainScaleS-2 \citep{pehle2022brainscales}, SpiNNaker-2, and Loihi-2 \citep{davies2021taking}. Recently, BPTT has been applied to SNNs, achieving higher accuracy compared to local learning rules \citep{wu2018spatio, wu2019direct}. Some works, like H2Learn \citep{liang2021h2learn} and SATA \citep{yin2022sata}, have designed specific architectures for BPTT learning in SNNs. 
In the future, the integration of learning rules will become increasingly important for exploring large and complex neuromorphic models in brain-inspired computing (BIC) chips. 

Another important type of BIC architecture is the compute-in-memory architecture, where in-core processing units and on-chip storage are physically integrated, performing synaptic integration matrix operations in synaptic memory. Compute-in-memory chips can be divided into two categories based on the materials: traditional or emerging memories. Traditional memories (such as SRAM, DRAM, and Flash) can be redesigned to support specific logical operations \citep{wu2020compact}. Their advantages include a mature ecosystem, easy simulation, and manufacturing. Emerging memories mainly refer to storage devices based on memristors. Synaptic weight storage, multiplication calculations, and presynaptic inputs are performed at the same crosspoint in the memristor, integrating computation and storage. 
\cite{zhang2021hybrid} designed a hybrid spiking neuron combining a memristor with simple digital circuits to enhance neuron functions. Further, they demonstrated a fully hardware spiking neural network with the hybrid neurons and memristive synapses for the first time, and achieved \textit{in-situ} Hebbian learning.
Brain-inspired computing hardware based on memristors involves multiple levels of material and architectural designs, which is currently still in a small-scale phase due to manufacturing process limitations.

Multiple types of brain-like chips have shown remarkable developments, demonstrating significant advantages in terms of biological simulation and low-power inference. However, they still face numerous challenges in practical applications. When it comes to handling high-level intelligence tasks, the superiority of brain-like chips compared to GPUs and ANN accelerators has not been fully established. Currently, to optimize their performance, some designs draw inspiration from ANN accelerators for improvements. It's worth noting that current brain-like chips do not yet support the training of large-scale SNNs and require special architectural designs to accommodate the training process for SNNs. To further support large-scale SNNs, it is necessary to enhance brain-like systems from both a software and hardware perspective in a more collaborative manner.

\subsection{Software and Hardware Interplay}
{The deployment of algorithms for SNNs onto neuromorphic chips typically requires certain software frameworks. 
The computational software frameworks mentioned in Section \ref{software} usually support simulations on mainstream CPUs and GPUs, without clear mention of deployment on neuromorphic chips. 
Meanwhile, among the previously mentioned neuromorphic chips in Section \ref{Hardware}, only about 27 percent of them are connected to application software packages \citep{schuman2022application}.
Typically, these application software packages contain model construction tools, simulators, and optimization tools. Model construction tools are used to define the structure and parameters of neural networks, including neuron types, connection patterns. Simulators are applied to simulate and debug neural network models on the chip. Optimization tools are adopted to train the network parameters and optimize its performance. 
Here are some typical examples. The Neurogrid chip is paired with the Neurogrid Software Framework \citep{benjamin2014neurogrid}, allowing users to specify neuronal models in the Python programming environment. 
The software framework for the BrainScaleS chip is the BrainScaleS-Software-Stack \citep{pehle2022brainscales}, which supports training neural networks on the chip using the PyTorch framework. 
The IBM TrueNorth chip typically utilizes a software framework called the TrueNorth Ecosystem, which is developed by the TrueNorth native Corelet language \citep{akopyan2015truenorth}. 
{IBM NorthPole \citep{modha2023ibm} is a brain-inspired memory-near-compute chip with a software development kit, but this chip does not emulate spiking communication.} 
Tianjic's software toolchain supports both ANN-to-SNN conversion and direct training for SNNs, and supports automatically transforming a pretrained model into an equivalent network that meets the Tianjic hardware constraints for non-spiking ANNs \citep{pei2019towards}. 
The latest Darwin3 builds a specialized instruction set architecture (ISA) \citep{ma2024darwin3}, which is close to machine code tailored for efficient neuromorphic computing. 
These software frameworks enable users to conveniently construct, simulate, and optimize neural network models on neuromorphic chips, facilitating efficient research and application development.
}

\section{Applications of Deep Spiking Neural Networks}\label{sec_app}
SNNs offer powerful computation capability due to their event-driven nature and temporal processing property. Theoretically, SNNs could be applied to any field where conventional deep neural networks (DNNs) are applied. As the training methods and programming frameworks of deep SNNs become more powerful, deep SNNs are increasingly drawing more attention and being applied to more fields, mainly including computer vision, reinforcement learning and autonomous robotics, biological visual system modeling, biological signal processing, natural language processing, equipment safety monitoring, and so on. It should be noted that this paper only lists some typical examples in recent years for some common application fields, not aiming to fully review all related studies. 

\subsection{Applications in Computer Vision}\label{sec_app_cv}
As traditional DNNs, the most common applications of SNNs lay in computer vision tasks. 
There are mainly two types of visual inputs for SNNs, i.e., RGB frames from traditional cameras or events from neuromorphic vision sensors. 
Neuromorphic vision sensors display great potential for computer vision tasks under high-speed and low-light conditions \citep{li2021recent}. SNNs are excellent candidates for processing neuromorphic signals due to their event-driven nature and energy-efficient computing. 

Recognition task plays an important role in the rapid progress of deep SNNs. SNNs are usually tested on both static datasets such as CIFAR10, CIFAR100, ImageNet, and neuromorphic datasets such as CIFAR10-DVS and DVS128-Gesture. Table \ref{tab:performance_on_datasets} lists the performances of some recently proposed architectures. 
Besides common recognition tasks, deep SNNs are increasingly applied to more computer vision tasks, including object detection/tracking, image denoising/generation, image/video reconstruction, video action recognition, image segmentation, and so on. 


\subsubsection{Object Detection and Object Tracking}\label{sec_app_cv_oDoT}
The first spike-based object detection model Spiking-YOLO was obtained through the ANN-to-SNN conversion method, achieving comparable performances to tiny-YOLO on PASCAL VOC and MS-COCO dataset with 3500 time steps \citep{kim2020spiking}. Later, a spike calibration (SpiCalib) method was proposed to reduce the time steps to hundreds \citep{li2022spike}. 
\cite{kugele2021hybrid} and \cite{cordone2022object} combined some spiking backbones with an SSD detection head for event cameras. 
\cite{su2023deep} proposed EMS-YOLO, the first directly trained deep SNNs for object detection, achieving comparable performance to the ANN counterpart while consuming less energy on both the frame-based MS-COCO dataset and the event-based Gen1 dataset in only 4 time steps. 

Considering that the Siamese networks have achieved remarkable performances in object tracking, SiamSNN was constructed by conversion to achieve short latency and low precision degradation on several benchmarks \citep{luo2022conversion}. Similarly, the directly trained Spiking SiamFC++ showed a small precision loss compared to the original SiamFC++ \citep{xiang2022spiking}. 
A spiking transformer network called STNet was developed for event-based single-object tracking, demonstrating competitive tracking accuracy and speed on three event-based datasets \citep{zhang2022spiking}. 
To process frames and events simultaneously, \cite{yang2019dashnet} proposed DashNet, achieving good tracking performance with a surprising tracking speed of 2083 FPS on neuromorphic chips. 

\subsubsection{Image Generation/Denoising and Image/Video Reconstruction}\label{sec_app_cv_imaG_imaR}
Generation tasks are increasingly explored in SNNs. 
\cite{comcsa2021spiking} introduced a directly trained spiking autoencoder to reconstruct images with high fidelity on MNIST and FMNIST. \cite{kamata2022fully} constructed a fully spiking variational autoencoder (FSVAE), generating images with competitive quality compared to conventional ANNs. \cite{liu2023spiking} proposed a Spiking-Diffusion model, outperforming the existing SNN-based generation models on several datasets. 
\cite{castagnetti2023spiden} developed an image denoising solution based on a directly trained spiking autoencoder, achieving a competitive signal-to-noise ratio on the Set12 dataset with significantly lower energy. 

Visual information reconstruction is important for neuromorphic vision sensors, because humans cannot directly perceive visual scenes from events. \cite{zhu2023revrec} provided a comprehensive review of visual reconstruction methods for events. 
\cite{duwek2021image} proposed a hybrid ANN-SNN model, accomplishing image reconstruction for simple scenes from N-MNIST and N-Caltech101 datasets. 
\cite{zhu2021neuspike} proposed an image reconstruction algorithm that combines DVS and Vidar signals, leveraging the high dynamic range of DVS to improve reconstruction effectiveness. Subsequently, they developed a deep SNN with an encoder-decoder structure for event-based video reconstruction, achieving performance comparable to ANN counterparts with only 0.05x energy consumption \citep{zhu2022event}.

\subsubsection{Others}\label{sec_app_cv_other}
Besides the above-mentioned tasks, deep SNNs are also applied in some other computer vision tasks, including video action recognition \citep{panda2018learning, chakraborty2023heterogeneous, wang2019temporal, zhang2022high}, image segmentation \citep{patel2021spiking, parameshwara2021spikems, kim2022beyond, zhang2023energy}, optical flow estimation \citep{lee2020spike, kosta2023adaptive, cuadrado2023optical}, depth prediction \citep{ranccon2022stereospike, wu2022mss, zhang2022spike}, point clouds processing \citep{zhou2020deep, ren2023spiking}, human pose tracking \citep{zou2023event}, lip-reading \citep{bulzomi2023end}, emotion/expression recognition \citep{Wang2022SpikingED, barchid2023spiking}, medical image classification \citep{shan2022detecting, qasim2023skin}, and so on.

\subsection{Applications in Other Fields}\label{sec_app_other}
Besides computer vision tasks, SNNs are showing gradually expanding application prospects in many fields, including reinforcement learning and autonomous robotics, biological visual system modeling, biological signal processing, natural language processing, equipment safety monitoring, and so on.

\subsubsection{Reinforcement Learning and Autonomous Robotics}\label{sec_app_other_rl}
As reinforcement learning (RL) is critical for the survival of humans and animals, there is increasing interest in applying brain-inspired SNNs to reinforcement learning. 
\cite{asgari2020digital} implemented an SNN with RL capability on hardware to learn stimulus-response associations, performing well in controlling a robot through a closed sensory-motor loop. 
\cite{chen2022deep} constructed the deep spiking Q-network (DSQN) to directly learn robust policies from high-dimensional sensory inputs using end-to-end deep RL, which was tested on 17 Atari games, outperforming the ANN-based deep Q-network (DQN) in most games. 
To reduce the latency of spiking RL, \cite{qin2023low} applied learnable matrix multiplication to encode and decode spikes.

Due to the good biological plausibility and high energy efficiency, SNNs have been applied to autonomous robotics for a long time, which is still a flourishing research direction, mainly including pattern generation (walk, trot, or run), motor control, and navigation (simultaneous localization and mapping, SLAM). 
\cite{yamazaki2022spiking} have already provided a good review of relevant studies, we do not go into more detail about this topic in this review which mainly focuses on deep SNNs. 

\subsubsection{Biological Visual System Modelling and Biological Signal Processing}\label{sec_app_other_bio}
ANNs play important roles in modeling biological visual pathways. However, SNNs are more biologically plausible models due to the use of temporal spike sequences. Therefore, several studies adopted SNNs to model the biological visual cortex. 
\cite{Huang2023DeepSN} applied deep SNNs to model the visual cortex for the first time, of which the similarity scores are higher than their ANN counterparts, showing SNNs' effectiveness. 
Further, they added a brain-inspired recurrent module into deep SNNs, outperforming the forward deep SNNs under natural movie stimuli \citep{huang2023deeprec}. 
\cite{zhang2023predicting} compared performances of deep SNNs and CNNs in the prediction of visual responses to naturalistic stimuli in three brain areas. 
\cite{ma2023temporal} presented a temporal conditioning spiking latent variable model to produce more realistic spike activities.

Due to the intrinsic dynamics, SNNs are also applied to process biological signals. 
\cite{xiong2021odor} proposed a convolutional SNN for odor recognition of electronic noses. 
\cite{feng2022building} applied the conversion method to obtain a 14-layer SNN model for ECG (electrocardiogram) classification. 
\cite{xiaoxue2023review} reviewed recent studies applying SNNs on signal classification and disease diagnosis based on biological signals including EEG, ECG, EMG, and so on.

\subsubsection{Others}\label{sec_app_other_other}
SNNs were also applied to natural language processing, equipment safety monitoring, semantic communication, multi-modal information processing, and so on. 
To ease the heavy energy cost of ANN-based large language models, some studies applied SNN-based architectures, including SpikBERT \citep{lv2023spikebert} and SpikeGPT \citep{zhu2023spikegpt}.
Applications regarding equipment safety monitoring mainly include battery health monitoring \citep{wang2023brain, wang2023bioinspired}, autonomous vehicle sensors fault diagnosis \citep{wang2023bioinspiredmem}, and bearing fault diagnosis \citep{xu2022deep}. 
Applications in semantic communication mainly tried to mitigate the limitation of transmission bandwidth \citep{wang2023spiking}.
Applications in multi-modal information processing currently show up in audio-visual zero-shot learning tasks \citep{li2023modality, li2023motion}.

{
\subsection{Discussion on SNN Applications}\label{sec_app_discuss}
Deep SNNs have achieved great success in many fields in recent years, but there still exist some limits needed to be addressed. 
Firstly, although many studies demonstrated that deep SNNs achieved comparable accuracy to their ANN counterparts on many tasks, they still lag behind conventional ANN SOTA, especially for large datasets like ImageNet, which asks for more endeavors. 
Secondly, many studies claimed that the proposed SNNs consumed much less energy compared to ANN counterparts, through calculating the number of addition operations, without considering the cost of other operations like data movement. Therefore, it is meaningful to deploy well-performed SNNs on neuromorphic chips or corresponding simulators to fully exploit the event-driven nature and measure the actual energy cost. 
Thirdly, as for applications requiring high processing speed and low power consumption, like robotics, it is promising to adopt neuromorphic vision/audio sensors and neuromorphic processing chips due to their event-driven nature, besides network pruning and weight quantization. Meanwhile, to fully exploiting the advantages of events, it deserves more efforts to explore how to directly process neuromorphic sensing events using SNNs, without converting events into frames as current studies usually do. 
Fourthly, as for transformer-based SNNs used in language or video processing, how to choose the input clip for one simulation step, to reconcile the temporal resolution of the input sequence and the simulation step of SNNs, is worth studying.
Last but not the least, as SNNs have an additional temporal dimension, how to achieve the speed-accuracy trade-off as humans is a problem worth of study. In other words, how to assign a suitable simulation duration or how to decide when to make a choice, are important questions to realize the balance between computation cost and prediction accuracy. 
}

\section{Future trends and conclusions}\label{Future trends}

This article provides an overview of the current developments in various theories and methods of deep SNNs, including relevant fundamentals, various spiking neuron models, advanced models, and architectures, booming software tools and hardware platforms, as well as applications in various fields. However, there are still many limitations and challenges.

(1) Currently, only a few aspects of the intelligent brains have been applied to instruct the construction and training of SNNs, lacking enough biological plausibility. Therefore, to improve SNNs' capability, it is necessary to introduce more types of spiking neurons, rich connection structures, multiscale local-global-cooperative learning rules, {system homeostasis,} etc., into SNNs to more accurately mimic the cognitive and intelligent characteristics emerging in the brains. {For example, it deserves more efforts to train SNNs with self-supervised learning, as children mainly receive unlabeled data during growth. Besides, the brain is actually a complex network, thus it is worthy of more efforts to study graph SNNs, although some attempts already exist.}

{(2) Recent neuroscience studies have found that astrocytes can naturally realize Transformer operations, which provides a new direction for the improvement of SNNs. In addition, astrocytes have the function of regulating neuronal firing activity and synaptic pruning, which provides ideas for the performance improvement and lightweight of SNNs in the future.}

{(3) }Information encoding methods and training algorithms for SNNs are mostly based on average firing rates, lacking the ability to represent temporal dynamics adequately. There should be more exploration of time-dependent information encoding strategies and corresponding training algorithms, to further enhance the spatiotemporal dynamic characteristics of SNNs and strengthen their temporal processing capability.

{(4) }The training of SNNs mainly employs time-dependent methods, like BPTT, which greatly increases the training cost, compared to conventional DNNs. Thus, there is a need to develop brain-like SNNs that can be trained in parallel, and dedicated software and hardware that support their computation, reducing training time and power consumption.

{(5) }As there are obstacles to conversion and interaction between different neuromorphic platforms, it is needed to establish a common standard to improve interoperability. Further, more brain-inspired principles or technologies should be incorporated into the design of neuromorphic systems, to enhance the computational performance of the chips, in terms of processing speed and energy efficiency. 

{(6) }Large-scale SNNs are mainly applied to classification tasks. Their potential in handling tasks that need to process continuous input streams, such as videos, languages, events from neuromorphic vision sensors, etc., has not been fully explored. Moreover, the introduction of various neuromorphic sensors and neuromorphic chips into autonomous robotics, cooperating with conventional sensors and processing chips, might be an efficient and effective way to achieve embodied intelligence. Further studies are needed to fully leverage the features and advantages of SNNs.

In summary, studies and applications of SNNs are growing rapidly, but there is still great potential to improve the effectiveness and efficiency of SNNs. Efforts should be made in multiple directions, including model architectures, training algorithms, software frameworks, and hardware platforms, to promote the coordinated progress of models, software, and hardware.

\section*{Conflict of Interest Statement}

The authors declare that the research was conducted in the absence of any commercial or financial relationships that could be construed as a potential conflict of interest.

\section*{Author Contributions}
Chenlin Zhou, Han Zhang, Liutao Yu, Yumin Ye: Writing - original draft, Writing - review \& editing. 
Zhaokun Zhou, Liwei Huang: Writing - review \& editing. 
Zhengyu Ma: Writing - review \& editing, Project administration, Supervision, Funding acquisition, Resources.
Xiaopeng Fan, Huihui Zhou, Yonghong Tian: Writing - review \& editing, Resources.
All authors contributed to the article and approved the submitted version.

\section*{Funding}
The study was funded by the National Natural Science Foundation of China under contracts No.62206141, No.62236009, No. 62332002, No. 62027804, No. 61825101, and the major key project of the Peng Cheng Laboratory (PCL2021A13). Computing support was provided by Pengcheng Cloudbrain.


\section*{Data Availability Statement}
The reviewed papers are collected in 
\href{https://github.com/zhouchenlin2096/Awesome-Spiking-Neural-Networks}{Awesome-Spiking-Neural-Networks}.


\bibliographystyle{Frontiers-Harvard} 
\bibliography{test}

\begin{thebibliography}{173}
\providecommand{\natexlab}[1]{#1}
\expandafter\ifx\csname urlstyle\endcsname\relax
  \providecommand{\doi}[1]{doi:\discretionary{}{}{}#1}\else
  \providecommand{\doi}{doi:\discretionary{}{}{}\begingroup \urlstyle{rm}\Url}\fi
\providecommand{\selectlanguage}[1]{\relax}
\providecommand{\bibAnnoteFile}[1]{%
  \IfFileExists{#1}{\begin{quotation}\noindent\textsc{Key:} #1\\
  \textsc{Annotation:}\ \input{#1}\end{quotation}}{}}
\providecommand{\bibAnnote}[2]{%
  \begin{quotation}\noindent\textsc{Key:} #1\\
  \textsc{Annotation:}\ #2\end{quotation}}

\bibitem[{Abadi et~al.(2016)Abadi, Barham, Chen, Chen, Davis, Dean et~al.}]{199317}
Abadi, M., Barham, P., Chen, J., Chen, Z., Davis, A., Dean, J., et~al. (2016).
\newblock Tensorflow: A system for {Large-Scale} machine learning.
\newblock In \emph{Symposium on Operating Systems Design and Implementation (OSDI)}. 265--283
\bibAnnoteFile{199317}

\bibitem[{Adrian and Zotterman(1926)}]{Impulse1926}
Adrian, E.~D. and Zotterman, Y. (1926).
\newblock The impulses produced by sensory nerve endings: Part 3. impulses set up by touch and pressure.
\newblock \emph{The Journal of physiology} 61, 465
\bibAnnoteFile{Impulse1926}

\bibitem[{Akopyan et~al.(2015)Akopyan, Sawada, Cassidy, Alvarez-Icaza, Arthur, Merolla et~al.}]{akopyan2015truenorth}
Akopyan, F., Sawada, J., Cassidy, A., Alvarez-Icaza, R., Arthur, J., Merolla, P., et~al. (2015).
\newblock Truenorth: Design and tool flow of a 65 mw 1 million neuron programmable neurosynaptic chip.
\newblock \emph{IEEE Transactions on Computer-aided Design of Integrated Circuits and Systems} 34, 1537--1557
\bibAnnoteFile{akopyan2015truenorth}

\bibitem[{Asgari et~al.(2020)Asgari, Maybodi, Kreiser, and Sandamirskaya}]{asgari2020digital}
Asgari, H., Maybodi, B. M.-N., Kreiser, R., and Sandamirskaya, Y. (2020).
\newblock Digital multiplier-less spiking neural network architecture of reinforcement learning in a context-dependent task.
\newblock \emph{IEEE Journal on Emerging and Selected Topics in Circuits and Systems} 10, 498--511
\bibAnnoteFile{asgari2020digital}

\bibitem[{Barchid et~al.(2023)Barchid, Allaert, Aissaoui, Mennesson, and Djeraba}]{barchid2023spiking}
Barchid, S., Allaert, B., Aissaoui, A., Mennesson, J., and Djeraba, C.~C. (2023).
\newblock Spiking-fer: spiking neural network for facial expression recognition with event cameras.
\newblock In \emph{International Conference on Content-based Multimedia Indexing (CBMI)}. 1--7
\bibAnnoteFile{barchid2023spiking}

\bibitem[{Bellec et~al.(2018)Bellec, Salaj, Subramoney, Legenstein, and Maass}]{Bellec2018LongSM}
Bellec, G., Salaj, D., Subramoney, A., Legenstein, R.~A., and Maass, W. (2018).
\newblock Long short-term memory and learning-to-learn in networks of spiking neurons.
\newblock In \emph{Advances in Neural Information Processing Systems (NeurIPS)}. vol.~31
\bibAnnoteFile{Bellec2018LongSM}

\bibitem[{Bellec et~al.(2020)Bellec, Scherr, Subramoney, Hajek, Salaj, Legenstein et~al.}]{bellec2020solution}
Bellec, G., Scherr, F., Subramoney, A., Hajek, E., Salaj, D., Legenstein, R., et~al. (2020).
\newblock A solution to the learning dilemma for recurrent networks of spiking neurons.
\newblock \emph{Nature communications} 11, 3625
\bibAnnoteFile{bellec2020solution}

\bibitem[{Benjamin et~al.(2014)Benjamin, Gao, McQuinn, Choudhary, Chandrasekaran, Bussat et~al.}]{benjamin2014neurogrid}
Benjamin, B.~V., Gao, P., McQuinn, E., Choudhary, S., Chandrasekaran, A.~R., Bussat, J.-M., et~al. (2014).
\newblock Neurogrid: A mixed-analog-digital multichip system for large-scale neural simulations.
\newblock \emph{Proceedings of the IEEE} 102, 699--716
\bibAnnoteFile{benjamin2014neurogrid}

\bibitem[{Bi and Poo(1998)}]{bi1998synaptic}
Bi, G.-q. and Poo, M.-m. (1998).
\newblock Synaptic modifications in cultured hippocampal neurons: dependence on spike timing, synaptic strength, and postsynaptic cell type.
\newblock \emph{Journal of Neuroscience} 18, 10464--10472
\bibAnnoteFile{bi1998synaptic}

\bibitem[{Bird and Polivoda(2021)}]{BPTT}
Bird, G.~M. and Polivoda, M.~E. (2021).
\newblock Backpropagation through time for networks with long-term dependencies.
\newblock \emph{arXiv} abs/2103.15589.
\newblock \doi{10.48550/arXiv.2103.15589}
\bibAnnoteFile{BPTT}

\bibitem[{Bohnstingl et~al.(2022)Bohnstingl, {\v{S}}urina, Fabre, Demira{\u{g}}, Frenkel, Payvand et~al.}]{bohnstingl2022biologically}
Bohnstingl, T., {\v{S}}urina, A., Fabre, M., Demira{\u{g}}, Y., Frenkel, C., Payvand, M., et~al. (2022).
\newblock Biologically-inspired training of spiking recurrent neural networks with neuromorphic hardware.
\newblock In \emph{2022 IEEE 4th International Conference on Artificial Intelligence Circuits and Systems (AICAS)} (IEEE), 218--221
\bibAnnoteFile{bohnstingl2022biologically}

\bibitem[{Bu et~al.(2022)Bu, Fang, Ding, Dai, Yu, and Huang}]{bu2021optimal}
Bu, T., Fang, W., Ding, J., Dai, P., Yu, Z., and Huang, T. (2022).
\newblock Optimal ann-snn conversion for high-accuracy and ultra-low-latency spiking neural networks.
\newblock In \emph{International Conference on Learning Representations (ICLR)}
\bibAnnoteFile{bu2021optimal}

\bibitem[{Bulzomi et~al.(2023)Bulzomi, Schweiker, Gruel, and Martinet}]{bulzomi2023end}
Bulzomi, H., Schweiker, M., Gruel, A., and Martinet, J. (2023).
\newblock End-to-end neuromorphic lip-reading.
\newblock In \emph{Proceedings of the IEEE/CVF Conference on Computer Vision and Pattern Recognition (CVPR)}. 4100--4107
\bibAnnoteFile{bulzomi2023end}

\bibitem[{Cao et~al.(2015)Cao, Chen, and Khosla}]{cao2015spiking}
Cao, Y., Chen, Y., and Khosla, D. (2015).
\newblock Spiking deep convolutional neural networks for energy-efficient object recognition.
\newblock \emph{International Journal of Computer Vision} 113, 54--66
\bibAnnoteFile{cao2015spiking}

\bibitem[{Carion et~al.(2020)Carion, Massa, Synnaeve, Usunier, Kirillov, and Zagoruyko}]{carion2020end}
Carion, N., Massa, F., Synnaeve, G., Usunier, N., Kirillov, A., and Zagoruyko, S. (2020).
\newblock End-to-end object detection with transformers.
\newblock In \emph{Proceedings of the European Conference on Computer Vision (ECCV)}. vol. 12346, 213--229
\bibAnnoteFile{carion2020end}

\bibitem[{Castagnetti et~al.(2023)Castagnetti, Pegatoquet, and Miramond}]{castagnetti2023spiden}
Castagnetti, A., Pegatoquet, A., and Miramond, B. (2023).
\newblock Spiden: deep spiking neural networks for efficient image denoising.
\newblock \emph{Frontiers in Neuroscience} 17, 1224457
\bibAnnoteFile{castagnetti2023spiden}

\bibitem[{Chakraborty and Mukhopadhyay(2023)}]{chakraborty2023heterogeneous}
Chakraborty, B. and Mukhopadhyay, S. (2023).
\newblock Heterogeneous recurrent spiking neural network for spatio-temporal classification.
\newblock \emph{Frontiers in Neuroscience} 17, 994517
\bibAnnoteFile{chakraborty2023heterogeneous}

\bibitem[{Chen et~al.(2022)Chen, Peng, Huang, and Tian}]{chen2022deep}
Chen, D., Peng, P., Huang, T., and Tian, Y. (2022).
\newblock Deep reinforcement learning with spiking q-learning.
\newblock \emph{arXiv} abs/2201.09754.
\newblock \doi{10.48550/arXiv.2201.09754}
\bibAnnoteFile{chen2022deep}

\bibitem[{Chen et~al.(2023)Chen, Peng, Li, and Tian}]{chen2023training}
Chen, G., Peng, P., Li, G., and Tian, Y. (2023).
\newblock Training full spike neural networks via auxiliary accumulation pathway.
\newblock \emph{arXiv} abs/2301.11929.
\newblock \doi{10.48550/arXiv.2301.11929}
\bibAnnoteFile{chen2023training}

\bibitem[{Chen et~al.(2018)Chen, Kumar, Sumbul, Knag, and Krishnamurthy}]{chen20184096}
Chen, G.~K., Kumar, R., Sumbul, H.~E., Knag, P.~C., and Krishnamurthy, R.~K. (2018).
\newblock A 4096-neuron 1m-synapse 3.8-pj/sop spiking neural network with on-chip stdp learning and sparse weights in 10-nm finfet cmos.
\newblock \emph{IEEE Journal of Solid-State Circuits} 54, 992--1002
\bibAnnoteFile{chen20184096}

\bibitem[{Com{\c{s}}a et~al.(2021)Com{\c{s}}a, Versari, Fischbacher, and Alakuijala}]{comcsa2021spiking}
Com{\c{s}}a, I.-M., Versari, L., Fischbacher, T., and Alakuijala, J. (2021).
\newblock Spiking autoencoders with temporal coding.
\newblock \emph{Frontiers in Neuroscience} 15, 712667
\bibAnnoteFile{comcsa2021spiking}

\bibitem[{Cordone et~al.(2022)Cordone, Miramond, and Thierion}]{cordone2022object}
Cordone, L., Miramond, B., and Thierion, P. (2022).
\newblock Object detection with spiking neural networks on automotive event data.
\newblock In \emph{International Joint Conference on Neural Networks (IJCNN)}. 1--8
\bibAnnoteFile{cordone2022object}

\bibitem[{Cuadrado et~al.(2023)Cuadrado, Ran{\c{c}}on, Cottereau, Barranco, and Masquelier}]{cuadrado2023optical}
Cuadrado, J., Ran{\c{c}}on, U., Cottereau, B.~R., Barranco, F., and Masquelier, T. (2023).
\newblock Optical flow estimation from event-based cameras and spiking neural networks.
\newblock \emph{Frontiers in Neuroscience} 17, 1160034
\bibAnnoteFile{cuadrado2023optical}

\bibitem[{Davies et~al.(2018)Davies, Srinivasa, Lin, Chinya, Cao, Choday et~al.}]{davies2018loihi}
Davies, M., Srinivasa, N., Lin, T.-H., Chinya, G., Cao, Y., Choday, S.~H., et~al. (2018).
\newblock Loihi: A neuromorphic manycore processor with on-chip learning.
\newblock \emph{IEEE Micro} 38, 82--99
\bibAnnoteFile{davies2018loihi}

\bibitem[{Davies et~al.(2021)}]{davies2021taking}
Davies, M. et~al. (2021).
\newblock Taking neuromorphic computing to the next level with loihi2.
\newblock \emph{Intel Labs’ Loihi} 2, 1--7
\bibAnnoteFile{davies2021taking}

\bibitem[{Deng et~al.(2022)Deng, Li, Zhang, and Gu}]{deng2021temporal}
Deng, S., Li, Y., Zhang, S., and Gu, S. (2022).
\newblock Temporal efficient training of spiking neural network via gradient re-weighting.
\newblock In \emph{International Conference on Learning Representations (ICLR)}
\bibAnnoteFile{deng2021temporal}

\bibitem[{Dosovitskiy et~al.(2021)Dosovitskiy, Beyer, Kolesnikov, Weissenborn, Zhai, Unterthiner et~al.}]{dosovitskiy2020image}
Dosovitskiy, A., Beyer, L., Kolesnikov, A., Weissenborn, D., Zhai, X., Unterthiner, T., et~al. (2021).
\newblock An image is worth 16x16 words: Transformers for image recognition at scale.
\newblock In \emph{International Conference on Learning Representations (ICLR)}
\bibAnnoteFile{dosovitskiy2020image}

\bibitem[{Duan et~al.(2022)Duan, Ding, Chen, Yu, and Huang}]{TEBN}
Duan, C., Ding, J., Chen, S., Yu, Z., and Huang, T. (2022).
\newblock Temporal effective batch normalization in spiking neural networks.
\newblock In \emph{Advances in Neural Information Processing Systems (NeurIPS)}. vol.~35, 34377--34390
\bibAnnoteFile{TEBN}

\bibitem[{Duwek et~al.(2021)Duwek, Shalumov, and Tsur}]{duwek2021image}
Duwek, H.~C., Shalumov, A., and Tsur, E.~E. (2021).
\newblock Image reconstruction from neuromorphic event cameras using laplacian-prediction and poisson integration with spiking and artificial neural networks.
\newblock In \emph{Proceedings of the IEEE/CVF Conference on Computer Vision and Pattern Recognition Workshops (CVPR Workshops)}. 1333--1341
\bibAnnoteFile{duwek2021image}

\bibitem[{Eshraghian et~al.(2023)Eshraghian, Ward, Neftci, Wang, Lenz, Dwivedi et~al.}]{eshraghian2023training}
Eshraghian, J.~K., Ward, M., Neftci, E.~O., Wang, X., Lenz, G., Dwivedi, G., et~al. (2023).
\newblock Training spiking neural networks using lessons from deep learning.
\newblock \emph{Proceedings of the IEEE} 111, 1016--1054
\bibAnnoteFile{eshraghian2023training}

\bibitem[{Fang et~al.(2023{\natexlab{a}})Fang, Chen, Ding, Yu, Masquelier, Chen et~al.}]{fang2023spikingjelly}
Fang, W., Chen, Y., Ding, J., Yu, Z., Masquelier, T., Chen, D., et~al. (2023{\natexlab{a}}).
\newblock Spikingjelly: An open-source machine learning infrastructure platform for spike-based intelligence.
\newblock \emph{Science Advances} 9, eadi1480
\bibAnnoteFile{fang2023spikingjelly}

\bibitem[{Fang et~al.(2021{\natexlab{a}})Fang, Yu, Chen, Huang, Masquelier, and Tian}]{fang2021deep}
Fang, W., Yu, Z., Chen, Y., Huang, T., Masquelier, T., and Tian, Y. (2021{\natexlab{a}}).
\newblock Deep residual learning in spiking neural networks.
\newblock In \emph{Advances in Neural Information Processing Systems (NeurIPS)}. vol.~34, 21056--21069
\bibAnnoteFile{fang2021deep}

\bibitem[{Fang et~al.(2021{\natexlab{b}})Fang, Yu, Chen, Masquelier, Huang, and Tian}]{fang2021incorporating}
Fang, W., Yu, Z., Chen, Y., Masquelier, T., Huang, T., and Tian, Y. (2021{\natexlab{b}}).
\newblock Incorporating learnable membrane time constant to enhance learning of spiking neural networks.
\newblock In \emph{Proceedings of the IEEE/CVF International Conference on Computer Vision (ICCV)}. 2661--2671
\bibAnnoteFile{fang2021incorporating}

\bibitem[{Fang et~al.(2023{\natexlab{b}})Fang, Yu, Zhou, Chen, Chen, Ma et~al.}]{fang2023parallel}
Fang, W., Yu, Z., Zhou, Z., Chen, D., Chen, Y., Ma, Z., et~al. (2023{\natexlab{b}}).
\newblock Parallel spiking neurons with high efficiency and ability to learn long-term dependencies.
\newblock In \emph{Advances in Neural Information Processing Systems (NeurIPS)}. vol.~36
\bibAnnoteFile{fang2023parallel}

\bibitem[{Feng et~al.(2022{\natexlab{a}})Feng, Liu, Tang, Ma, and Pan}]{mlf}
Feng, L., Liu, Q., Tang, H., Ma, D., and Pan, G. (2022{\natexlab{a}}).
\newblock Multi-level firing with spiking ds-resnet: Enabling better and deeper directly-trained spiking neural networks.
\newblock In \emph{Proceedings of the Thirty-First International Joint Conference on Artificial Intelligence (IJCAI))}. 2471--2477
\bibAnnoteFile{mlf}

\bibitem[{Feng et~al.(2022{\natexlab{b}})Feng, Geng, Chu, Fu, and Hong}]{feng2022building}
Feng, Y., Geng, S., Chu, J., Fu, Z., and Hong, S. (2022{\natexlab{b}}).
\newblock Building and training a deep spiking neural network for ecg classification.
\newblock \emph{Biomedical Signal Processing and Control} 77, 103749
\bibAnnoteFile{feng2022building}

\bibitem[{Frenkel et~al.(2018)Frenkel, Lefebvre, Legat, and Bol}]{frenkel20180}
Frenkel, C., Lefebvre, M., Legat, J.-D., and Bol, D. (2018).
\newblock A 0.086-$mm^2 $ 12.7-pj/sop 64k-synapse 256-neuron online-learning digital spiking neuromorphic processor in 28-nm cmos.
\newblock \emph{IEEE Transactions on Biomedical Circuits and Systems} 13, 145--158
\bibAnnoteFile{frenkel20180}

\bibitem[{Frenkel et~al.(2019)Frenkel, Legat, and Bol}]{frenkel2019morphic}
Frenkel, C., Legat, J.-D., and Bol, D. (2019).
\newblock Morphic: A 65-nm 738k-synapse/$mm^2$ quad-core binary-weight digital neuromorphic processor with stochastic spike-driven online learning.
\newblock \emph{IEEE Transactions on Biomedical Circuits and Systems} 13, 999--1010
\bibAnnoteFile{frenkel2019morphic}

\bibitem[{Goodman and Brette(2009)}]{goodman2009brian}
Goodman, D.~F. and Brette, R. (2009).
\newblock The brian simulator.
\newblock \emph{Frontiers in Neuroscience} 3, 643
\bibAnnoteFile{goodman2009brian}

\bibitem[{Guo et~al.(2021{\natexlab{a}})Guo, Bethge, Yang, Zhong, Ning, Meinel et~al.}]{guo2021boolnet}
Guo, N., Bethge, J., Yang, H., Zhong, K., Ning, X., Meinel, C., et~al. (2021{\natexlab{a}}).
\newblock Boolnet: minimizing the energy consumption of binary neural networks.
\newblock \emph{arXiv} abs/2106.06991.
\newblock \doi{10.48550/arXiv.2106.06991}
\bibAnnoteFile{guo2021boolnet}

\bibitem[{Guo et~al.(2021{\natexlab{b}})Guo, Fouda, Eltawil, and Salama}]{neuralcodingzongshu}
Guo, W., Fouda, M.~E., Eltawil, A.~M., and Salama, K.~N. (2021{\natexlab{b}}).
\newblock Neural coding in spiking neural networks: A comparative study for robust neuromorphic systems.
\newblock \emph{Frontiers in Neuroscience} 15, 638474
\bibAnnoteFile{neuralcodingzongshu}

\bibitem[{Guo et~al.(2022{\natexlab{a}})Guo, Chen, Zhang, Liu, Wang, Huang et~al.}]{IMLoss}
Guo, Y., Chen, Y., Zhang, L., Liu, X., Wang, Y., Huang, X., et~al. (2022{\natexlab{a}}).
\newblock Im-loss: Information maximization loss for spiking neural networks.
\newblock In \emph{Advances in Neural Information Processing Systems (NeurIPS)}. vol.~35, 156--166
\bibAnnoteFile{IMLoss}

\bibitem[{Guo et~al.(2023{\natexlab{a}})Guo, Liu, Chen, Zhang, Peng, Zhang et~al.}]{guo2023rmp}
Guo, Y., Liu, X., Chen, Y., Zhang, L., Peng, W., Zhang, Y., et~al. (2023{\natexlab{a}}).
\newblock Rmp-loss: Regularizing membrane potential distribution for spiking neural networks.
\newblock In \emph{Proceedings of the IEEE/CVF International Conference on Computer Vision (ICCV)}. 17391--17401
\bibAnnoteFile{guo2023rmp}

\bibitem[{Guo et~al.(2022{\natexlab{b}})Guo, Tong, Chen, Zhang, Liu, Ma et~al.}]{RecDis-SNN}
Guo, Y., Tong, X., Chen, Y., Zhang, L., Liu, X., Ma, Z., et~al. (2022{\natexlab{b}}).
\newblock Recdis-snn: Rectifying membrane potential distribution for directly training spiking neural networks.
\newblock In \emph{Proceedings of the IEEE/CVF Conference on Computer Vision and Pattern Recognition (CVPR)}. 326--335
\bibAnnoteFile{RecDis-SNN}

\bibitem[{Guo et~al.(2023{\natexlab{b}})Guo, Zhang, Chen, Peng, Liu, Zhang et~al.}]{guo2023membrane}
Guo, Y., Zhang, Y., Chen, Y., Peng, W., Liu, X., Zhang, L., et~al. (2023{\natexlab{b}}).
\newblock Membrane potential batch normalization for spiking neural networks.
\newblock In \emph{Proceedings of the IEEE/CVF International Conference on Computer Vision (ICCV)}. 19420--19430
\bibAnnoteFile{guo2023membrane}

\bibitem[{Hamilton et~al.(2019)Hamilton, Mintz, and Schuman}]{hamilton2019spike}
Hamilton, K.~E., Mintz, T.~M., and Schuman, C.~D. (2019).
\newblock Spike-based primitives for graph algorithms.
\newblock \emph{arXiv preprint arXiv:1903.10574}
\bibAnnoteFile{hamilton2019spike}

\bibitem[{Hazan et~al.(2018)Hazan, Saunders, Khan, Patel, Sanghavi, Siegelmann et~al.}]{hazan2018bindsnet}
Hazan, H., Saunders, D.~J., Khan, H., Patel, D., Sanghavi, D.~T., Siegelmann, H.~T., et~al. (2018).
\newblock Bindsnet: A machine learning-oriented spiking neural networks library in python.
\newblock \emph{Frontiers in Neuroinformatics} 12, 89
\bibAnnoteFile{hazan2018bindsnet}

\bibitem[{Hines and Carnevale(1997)}]{hines1997neuron}
Hines, M.~L. and Carnevale, N.~T. (1997).
\newblock The neuron simulation environment.
\newblock \emph{Neural Computation} 9, 1179--1209
\bibAnnoteFile{hines1997neuron}

\bibitem[{Hu et~al.(2021{\natexlab{a}})Hu, Tang, and Pan}]{hu2021spiking}
Hu, Y., Tang, H., and Pan, G. (2021{\natexlab{a}}).
\newblock Spiking deep residual networks.
\newblock \emph{IEEE Transactions on Neural Networks and Learning Systems} 34, 5200--5205
\bibAnnoteFile{hu2021spiking}

\bibitem[{Hu et~al.(2021{\natexlab{b}})Hu, Wu, Deng, and Li}]{hu2021advancing}
Hu, Y., Wu, Y., Deng, L., and Li, G. (2021{\natexlab{b}}).
\newblock Advancing residual learning towards powerful deep spiking neural networks.
\newblock \emph{arXiv} abs/2112.08954.
\newblock \doi{10.48550/arXiv.2112.08954}
\bibAnnoteFile{hu2021advancing}

\bibitem[{Huang et~al.(2023{\natexlab{a}})Huang, Ma, Yu, Zhou, and Tian}]{Huang2023DeepSN}
Huang, L., Ma, Z., Yu, L., Zhou, H., and Tian, Y. (2023{\natexlab{a}}).
\newblock Deep spiking neural networks with high representation similarity model visual pathways of macaque and mouse.
\newblock In \emph{Proceedings of the AAAI Conference on Artificial Intelligence (AAAI)}. 31--39
\bibAnnoteFile{Huang2023DeepSN}

\bibitem[{Huang et~al.(2023{\natexlab{b}})Huang, Ma, Zhou, and Tian}]{huang2023deeprec}
Huang, L., Ma, Z., Zhou, H., and Tian, Y. (2023{\natexlab{b}}).
\newblock Deep recurrent spiking neural networks capture both static and dynamic representations of the visual cortex under movie stimuli.
\newblock \emph{arXiv} abs/2306.01354.
\newblock \doi{10.48550/arXiv.2306.01354}
\bibAnnoteFile{huang2023deeprec}

\bibitem[{Hunsberger and Eliasmith(2015)}]{hunsberger2015spiking}
Hunsberger, E. and Eliasmith, C. (2015).
\newblock Spiking deep networks with lif neurons.
\newblock \emph{arXiv} abs/1510.08829.
\newblock \doi{10.48550/arXiv.1510.08829}
\bibAnnoteFile{hunsberger2015spiking}

\bibitem[{Jiang and Zhang(2023)}]{Jiang2023KLIFAO}
Jiang, C. and Zhang, Y. (2023).
\newblock Klif: An optimized spiking neuron unit for tuning surrogate gradient slope and membrane potential.
\newblock \emph{arXiv} abs/2302.09238.
\newblock \doi{10.48550/arXiv.2302.09238}
\bibAnnoteFile{Jiang2023KLIFAO}

\bibitem[{Kamata et~al.(2022)Kamata, Mukuta, and Harada}]{kamata2022fully}
Kamata, H., Mukuta, Y., and Harada, T. (2022).
\newblock Fully spiking variational autoencoder.
\newblock In \emph{Proceedings of the AAAI Conference on Artificial Intelligence (AAAI)}. 7059--7067
\bibAnnoteFile{kamata2022fully}

\bibitem[{Kim et~al.(2020)Kim, Park, Na, and Yoon}]{kim2020spiking}
Kim, S., Park, S., Na, B., and Yoon, S. (2020).
\newblock Spiking-yolo: spiking neural network for energy-efficient object detection.
\newblock In \emph{Proceedings of the AAAI Conference on Artificial Intelligence (AAAI)}. 11270--11277
\bibAnnoteFile{kim2020spiking}

\bibitem[{Kim et~al.(2022)Kim, Chough, and Panda}]{kim2022beyond}
Kim, Y., Chough, J., and Panda, P. (2022).
\newblock Beyond classification: Directly training spiking neural networks for semantic segmentation.
\newblock \emph{Neuromorphic Computing and Engineering} 2, 044015
\bibAnnoteFile{kim2022beyond}

\bibitem[{Kim and Panda(2021)}]{BNTT}
Kim, Y. and Panda, P. (2021).
\newblock Revisiting batch normalization for training low-latency deep spiking neural networks from scratch.
\newblock \emph{Frontiers in Neuroscience} 15, 773954
\bibAnnoteFile{BNTT}

\bibitem[{Kosta and Roy(2023)}]{kosta2023adaptive}
Kosta, A.~K. and Roy, K. (2023).
\newblock Adaptive-spikenet: event-based optical flow estimation using spiking neural networks with learnable neuronal dynamics.
\newblock In \emph{International Conference on Robotics and Automation (ICRA)}. 6021--6027
\bibAnnoteFile{kosta2023adaptive}

\bibitem[{Kuang et~al.(2021)Kuang, Cui, Zhong, Liu, Zou, Dai et~al.}]{kuang202164k}
Kuang, Y., Cui, X., Zhong, Y., Liu, K., Zou, C., Dai, Z., et~al. (2021).
\newblock A 64k-neuron 64m-1b-synapse 2.64 pj/sop neuromorphic chip with all memory on chip for spike-based models in 65nm cmos.
\newblock \emph{IEEE Transactions on Circuits and Systems II: Express Briefs} 68, 2655--2659
\bibAnnoteFile{kuang202164k}

\bibitem[{Kugele et~al.(2021)Kugele, Pfeil, Pfeiffer, and Chicca}]{kugele2021hybrid}
Kugele, A., Pfeil, T., Pfeiffer, M., and Chicca, E. (2021).
\newblock Hybrid snn-ann: Energy-efficient classification and object detection for event-based vision.
\newblock In \emph{DAGM German Conference on Pattern Recognition (GCPR)}. 297--312
\bibAnnoteFile{kugele2021hybrid}

\bibitem[{Lee et~al.(2020{\natexlab{a}})Lee, Kosta, Zhu, Chaney, Daniilidis, and Roy}]{lee2020spike}
Lee, C., Kosta, A.~K., Zhu, A.~Z., Chaney, K., Daniilidis, K., and Roy, K. (2020{\natexlab{a}}).
\newblock Spike-flownet: event-based optical flow estimation with energy-efficient hybrid neural networks.
\newblock In \emph{Proceedings of the European Conference on Computer Vision (ECCV)}. vol. 12374, 366--382
\bibAnnoteFile{lee2020spike}

\bibitem[{Lee et~al.(2020{\natexlab{b}})Lee, Sarwar, Panda, Srinivasan, and Roy}]{lee2020enabling}
Lee, C., Sarwar, S.~S., Panda, P., Srinivasan, G., and Roy, K. (2020{\natexlab{b}}).
\newblock Enabling spike-based backpropagation for training deep neural network architectures.
\newblock \emph{Frontiers in Neuroscience} 14, 119
\bibAnnoteFile{lee2020enabling}

\bibitem[{Lee et~al.(2016)Lee, Delbruck, and Pfeiffer}]{lee2016training}
Lee, J.~H., Delbruck, T., and Pfeiffer, M. (2016).
\newblock Training deep spiking neural networks using backpropagation.
\newblock \emph{Frontiers in Neuroscience} 10, 508
\bibAnnoteFile{lee2016training}

\bibitem[{Li and Tian(2021)}]{li2021recent}
Li, J. and Tian, Y. (2021).
\newblock Recent advances in neuromorphic vision sensors: A survey.
\newblock \emph{Chinese Journal of Computers} 44, 1258--1286
\bibAnnoteFile{li2021recent}

\bibitem[{Li et~al.(2023{\natexlab{a}})Li, Ma, Deng, Man, and Fan}]{li2023modality}
Li, W., Ma, Z., Deng, L.-J., Man, H., and Fan, X. (2023{\natexlab{a}}).
\newblock Modality-fusion spiking transformer network for audio-visual zero-shot learning.
\newblock In \emph{International Conference on Multimedia and Expo (ICME)}. 426--431
\bibAnnoteFile{li2023modality}

\bibitem[{Li et~al.(2023{\natexlab{b}})Li, Zhao, Ma, Wang, Fan, and Tian}]{li2023motion}
Li, W., Zhao, X.-L., Ma, Z., Wang, X., Fan, X., and Tian, Y. (2023{\natexlab{b}}).
\newblock Motion-decoupled spiking transformer for audio-visual zero-shot learning.
\newblock In \emph{Proceedings of the International Conference on Multimedia (MM)}. 3994--4002
\bibAnnoteFile{li2023motion}

\bibitem[{Li et~al.(2023{\natexlab{c}})Li, Zhang, Yi, Liu, Wang, Zhang et~al.}]{xiaoxue2023review}
Li, X., Zhang, X., Yi, X., Liu, D., Wang, H., Zhang, B., et~al. (2023{\natexlab{c}}).
\newblock Review of medical data analysis based on spiking neural networks.
\newblock \emph{Procedia Computer Science} 221, 1527--1538
\bibAnnoteFile{xiaoxue2023review}

\bibitem[{Li et~al.(2021)Li, Guo, Zhang, Deng, Hai, and Gu}]{Dspike}
Li, Y., Guo, Y., Zhang, S., Deng, S., Hai, Y., and Gu, S. (2021).
\newblock Differentiable spike: Rethinking gradient-descent for training spiking neural networks.
\newblock In \emph{Advances in Neural Information Processing Systems (NeurIPS)}. vol.~34, 23426--23439
\bibAnnoteFile{Dspike}

\bibitem[{Li et~al.(2022)Li, He, Dong, Kong, and Zeng}]{li2022spike}
Li, Y., He, X., Dong, Y., Kong, Q., and Zeng, Y. (2022).
\newblock Spike calibration: Fast and accurate conversion of spiking neural network for object detection and segmentation.
\newblock \emph{arXiv} abs/2207.02702.
\newblock \doi{10.48550/arXiv.2207.02702}
\bibAnnoteFile{li2022spike}

\bibitem[{Lian et~al.(2023)Lian, Shen, Liu, Wang, Yan, and Tang}]{LearnableSG}
Lian, S., Shen, J., Liu, Q., Wang, Z., Yan, R., and Tang, H. (2023).
\newblock Learnable surrogate gradient for direct training spiking neural networks.
\newblock In \emph{Proceedings of the Thirty-Second International Joint Conference on Artificial Intelligence (IJCAI)}. 3002--3010
\bibAnnoteFile{LearnableSG}

\bibitem[{Liang et~al.(2022)Liang, Li, Wang, Zhang, Yue, Li et~al.}]{patel2021spiking}
Liang, J., Li, R., Wang, C., Zhang, R., Yue, K., Li, W., et~al. (2022).
\newblock A spiking neural network based on retinal ganglion cells for automatic burn image segmentation.
\newblock \emph{Entropy} 24, 1526
\bibAnnoteFile{patel2021spiking}

\bibitem[{Liang et~al.(2021)Liang, Qu, Chen, Tu, Wu, Deng et~al.}]{liang2021h2learn}
Liang, L., Qu, Z., Chen, Z., Tu, F., Wu, Y., Deng, L., et~al. (2021).
\newblock H2learn: High-efficiency learning accelerator for high-accuracy spiking neural networks.
\newblock \emph{IEEE Transactions on Computer-Aided Design of Integrated Circuits and Systems} 41, 4782--4796
\bibAnnoteFile{liang2021h2learn}

\bibitem[{Liu et~al.(2023)Liu, Wen, and Chen}]{liu2023spiking}
Liu, M., Wen, R., and Chen, H. (2023).
\newblock Spiking-diffusion: Vector quantized discrete diffusion model with spiking neural networks.
\newblock \emph{arXiv} abs/2308.10187.
\newblock \doi{10.48550/arXiv.2308.10187}
\bibAnnoteFile{liu2023spiking}

\bibitem[{Liu et~al.(2021)Liu, Lin, Cao, Hu, Wei, Zhang et~al.}]{liu2021swin}
Liu, Z., Lin, Y., Cao, Y., Hu, H., Wei, Y., Zhang, Z., et~al. (2021).
\newblock Swin transformer: Hierarchical vision transformer using shifted windows.
\newblock In \emph{Proceedings of the IEEE/CVF International Conference on Computer Vision (ICCV)}. 10012--10022
\bibAnnoteFile{liu2021swin}

\bibitem[{Liu et~al.(2020)Liu, Shen, Savvides, and Cheng}]{liu2020reactnet}
Liu, Z., Shen, Z., Savvides, M., and Cheng, K.-T. (2020).
\newblock Reactnet: Towards precise binary neural network with generalized activation functions.
\newblock In \emph{Proceedings of the European Conference on Computer Vision (ECCV)}. vol. 12359, 143--159
\bibAnnoteFile{liu2020reactnet}

\bibitem[{Liu et~al.(2018)Liu, Wu, Luo, Yang, Liu, and Cheng}]{liu2018bi}
Liu, Z., Wu, B., Luo, W., Yang, X., Liu, W., and Cheng, K.-T. (2018).
\newblock Bi-real net: Enhancing the performance of 1-bit cnns with improved representational capability and advanced training algorithm.
\newblock In \emph{Proceedings of the European Conference on Computer Vision (ECCV)}. vol. 11219, 747--763
\bibAnnoteFile{liu2018bi}

\bibitem[{Luo et~al.(2022)Luo, Shen, Cao, Wang, Feng, and Tan}]{luo2022conversion}
Luo, Y., Shen, H., Cao, X., Wang, T., Feng, Q., and Tan, Z. (2022).
\newblock Conversion of siamese networks to spiking neural networks for energy-efficient object tracking.
\newblock \emph{Neural Computing and Applications} 34, 9967--9982
\bibAnnoteFile{luo2022conversion}

\bibitem[{Lv et~al.(2023)Lv, Li, Xu, Gu, Ling, Zhang et~al.}]{lv2023spikebert}
Lv, C., Li, T., Xu, J., Gu, C., Ling, Z., Zhang, C., et~al. (2023).
\newblock Spikebert: A language spikformer trained with two-stage knowledge distillation from bert.
\newblock \emph{arXiv} abs/2308.15122.
\newblock \doi{10.48550/arXiv.2308.15122}
\bibAnnoteFile{lv2023spikebert}

\bibitem[{Ma et~al.(2024)Ma, Jin, Sun, Li, Wu, Hu et~al.}]{ma2024darwin3}
Ma, D., Jin, X., Sun, S., Li, Y., Wu, X., Hu, Y., et~al. (2024).
\newblock Darwin3: a large-scale neuromorphic chip with a novel isa and on-chip learning.
\newblock \emph{National Science Review} 11, nwae102
\bibAnnoteFile{ma2024darwin3}

\bibitem[{Ma et~al.(2017)Ma, Shen, Gu, Zhang, Zhu, Xu et~al.}]{ma2017darwin}
Ma, D., Shen, J., Gu, Z., Zhang, M., Zhu, X., Xu, X., et~al. (2017).
\newblock Darwin: A neuromorphic hardware co-processor based on spiking neural networks.
\newblock \emph{Journal of Systems Architecture} 77, 43--51
\bibAnnoteFile{ma2017darwin}

\bibitem[{Ma et~al.(2023)Ma, Jiang, Yan, and Tang}]{ma2023temporal}
Ma, G., Jiang, R., Yan, R., and Tang, H. (2023).
\newblock Temporal conditioning spiking latent variable models of the neural response to natural visual scenes.
\newblock In \emph{Advances in Neural Information Processing Systems (NeurIPS)}. vol.~36
\bibAnnoteFile{ma2023temporal}

\bibitem[{Maass(1997)}]{maass1997networks}
Maass, W. (1997).
\newblock Networks of spiking neurons: the third generation of neural network models.
\newblock \emph{Neural Networks} 10, 1659--1671
\bibAnnoteFile{maass1997networks}

\bibitem[{Meng et~al.(2022)Meng, Xiao, Yan, Wang, Lin, and Luo}]{meng2022training}
Meng, Q., Xiao, M., Yan, S., Wang, Y., Lin, Z., and Luo, Z.-Q. (2022).
\newblock Training high-performance low-latency spiking neural networks by differentiation on spike representation.
\newblock In \emph{Proceedings of the IEEE/CVF Conference on Computer Vision and Pattern Recognition (CVPR)}. 12444--12453
\bibAnnoteFile{meng2022training}

\bibitem[{Meng et~al.(2023)Meng, Xiao, Yan, Wang, Lin, and Luo}]{SLTT}
Meng, Q., Xiao, M., Yan, S., Wang, Y., Lin, Z., and Luo, Z.-Q. (2023).
\newblock Towards memory- and time-efficient backpropagation for training spiking neural networks.
\newblock In \emph{2023 IEEE/CVF International Conference on Computer Vision (ICCV)}. 6143--6153.
\newblock \doi{10.1109/ICCV51070.2023.00567}
\bibAnnoteFile{SLTT}

\bibitem[{Modha et~al.(2023)Modha, Akopyan, Andreopoulos, Appuswamy, Arthur, Cassidy et~al.}]{modha2023ibm}
Modha, D.~S., Akopyan, F., Andreopoulos, A., Appuswamy, R., Arthur, J.~V., Cassidy, A.~S., et~al. (2023).
\newblock Ibm northpole neural inference machine.
\newblock In \emph{2023 IEEE Hot Chips 35 Symposium (HCS)} (IEEE Computer Society), 1--58
\bibAnnoteFile{modha2023ibm}

\bibitem[{Moradi et~al.(2017)Moradi, Qiao, Stefanini, and Indiveri}]{moradi2017scalable}
Moradi, S., Qiao, N., Stefanini, F., and Indiveri, G. (2017).
\newblock A scalable multicore architecture with heterogeneous memory structures for dynamic neuromorphic asynchronous processors (dynaps).
\newblock \emph{IEEE Transactions on Biomedical Circuits and Systems} 12, 106--122
\bibAnnoteFile{moradi2017scalable}

\bibitem[{Mozafari et~al.(2019)Mozafari, Ganjtabesh, Nowzari-Dalini, and Masquelier}]{mozafari2019spyketorch}
Mozafari, M., Ganjtabesh, M., Nowzari-Dalini, A., and Masquelier, T. (2019).
\newblock Spyketorch: Efficient simulation of convolutional spiking neural networks with at most one spike per neuron.
\newblock \emph{Frontiers in Neuroscience} 13, 625
\bibAnnoteFile{mozafari2019spyketorch}

\bibitem[{Neftci et~al.(2019)Neftci, Mostafa, and Zenke}]{neftci2019surrogate}
Neftci, E.~O., Mostafa, H., and Zenke, F. (2019).
\newblock Surrogate gradient learning in spiking neural networks: Bringing the power of gradient-based optimization to spiking neural networks.
\newblock \emph{IEEE Signal Processing Magazine} 36, 51--63
\bibAnnoteFile{neftci2019surrogate}

\bibitem[{Painkras et~al.(2013)Painkras, Plana, Garside, Temple, Galluppi, Patterson et~al.}]{painkras2013spinnaker}
Painkras, E., Plana, L.~A., Garside, J., Temple, S., Galluppi, F., Patterson, C., et~al. (2013).
\newblock Spinnaker: A 1-w 18-core system-on-chip for massively-parallel neural network simulation.
\newblock \emph{IEEE Journal of Solid-State Circuits} 48, 1943--1953
\bibAnnoteFile{painkras2013spinnaker}

\bibitem[{Panda and Srinivasa(2018)}]{panda2018learning}
Panda, P. and Srinivasa, N. (2018).
\newblock Learning to recognize actions from limited training examples using a recurrent spiking neural model.
\newblock \emph{Frontiers in Neuroscience} 12, 126
\bibAnnoteFile{panda2018learning}

\bibitem[{Parameshwara et~al.(2021)Parameshwara, Li, Ferm{\"u}ller, Sanket, Evanusa, and Aloimonos}]{parameshwara2021spikems}
Parameshwara, C.~M., Li, S., Ferm{\"u}ller, C., Sanket, N.~J., Evanusa, M.~S., and Aloimonos, Y. (2021).
\newblock Spikems: Deep spiking neural network for motion segmentation.
\newblock In \emph{International Conference on Intelligent Robots and Systems (IROS)}. 3414--3420
\bibAnnoteFile{parameshwara2021spikems}

\bibitem[{Park et~al.(2020)Park, Kim, Na, and Yoon}]{t2fsnn}
Park, S., Kim, S., Na, B., and Yoon, S. (2020).
\newblock T2fsnn: Deep spiking neural networks with time-to-first-spike coding.
\newblock In \emph{ACM/IEEE Design Automation Conference (DAC)}. 1--6
\bibAnnoteFile{t2fsnn}

\bibitem[{Paszke et~al.(2019)Paszke, Gross, Massa, Lerer, Bradbury, Chanan et~al.}]{paszke2019pytorch}
Paszke, A., Gross, S., Massa, F., Lerer, A., Bradbury, J., Chanan, G., et~al. (2019).
\newblock Pytorch: An imperative style, high-performance deep learning library.
\newblock In \emph{Advances in Neural Information Processing Systems (NeurIPS)}. vol.~32
\bibAnnoteFile{paszke2019pytorch}

\bibitem[{Pehle et~al.(2022)Pehle, Billaudelle, Cramer, Kaiser, Schreiber, Stradmann et~al.}]{pehle2022brainscales}
Pehle, C., Billaudelle, S., Cramer, B., Kaiser, J., Schreiber, K., Stradmann, Y., et~al. (2022).
\newblock The brainscales-2 accelerated neuromorphic system with hybrid plasticity.
\newblock \emph{Frontiers in Neuroscience} 16, 795876
\bibAnnoteFile{pehle2022brainscales}

\bibitem[{Pei et~al.(2019)Pei, Deng, Song, Zhao, Zhang, Wu et~al.}]{pei2019towards}
Pei, J., Deng, L., Song, S., Zhao, M., Zhang, Y., Wu, S., et~al. (2019).
\newblock Towards artificial general intelligence with hybrid tianjic chip architecture.
\newblock \emph{Nature} 572, 106--111
\bibAnnoteFile{pei2019towards}

\bibitem[{Qasim~Gilani et~al.(2023)Qasim~Gilani, Syed, Umair, and Marques}]{qasim2023skin}
Qasim~Gilani, S., Syed, T., Umair, M., and Marques, O. (2023).
\newblock Skin cancer classification using deep spiking neural network.
\newblock \emph{Journal of Digital Imaging} 36, 1137–1147
\bibAnnoteFile{qasim2023skin}

\bibitem[{Qiao et~al.(2015)Qiao, Mostafa, adi, Osswald, Stefanini, Sumislawska et~al.}]{qiao2015reconfigurable}
Qiao, N., Mostafa, H., adi, F., Osswald, M., Stefanini, F., Sumislawska, D., et~al. (2015).
\newblock A reconfigurable on-line learning spiking neuromorphic processor comprising 256 neurons and 128k synapses.
\newblock \emph{Frontiers in Neuroscience} 9, 141
\bibAnnoteFile{qiao2015reconfigurable}

\bibitem[{Qin et~al.(2023)Qin, Yan, and Tang}]{qin2023low}
Qin, L., Yan, R., and Tang, H. (2023).
\newblock A low latency adaptive coding spike framework for deep reinforcement learning.
\newblock In \emph{Proceedings of the Thirty-Second International Joint Conference on Artificial Intelligence (IJCAI)}. 3049--3057
\bibAnnoteFile{qin2023low}

\bibitem[{Ran{\c{c}}on et~al.(2022)Ran{\c{c}}on, Cuadrado-Anibarro, Cottereau, and Masquelier}]{ranccon2022stereospike}
Ran{\c{c}}on, U., Cuadrado-Anibarro, J., Cottereau, B.~R., and Masquelier, T. (2022).
\newblock Stereospike: Depth learning with a spiking neural network.
\newblock \emph{IEEE Access} 10, 127428--127439
\bibAnnoteFile{ranccon2022stereospike}

\bibitem[{Rasmussen(2019)}]{rasmussen2019nengodl}
Rasmussen, D. (2019).
\newblock Nengodl: Combining deep learning and neuromorphic modelling methods.
\newblock \emph{Neuroinformatics} 17, 611--628
\bibAnnoteFile{rasmussen2019nengodl}

\bibitem[{Rathi and Roy(2023)}]{diet-snn-tnnls}
Rathi, N. and Roy, K. (2023).
\newblock Diet-snn: A low-latency spiking neural network with direct input encoding and leakage and threshold optimization.
\newblock \emph{IEEE Transactions on Neural Networks and Learning Systems} 34, 3174--3182
\bibAnnoteFile{diet-snn-tnnls}

\bibitem[{Rathi et~al.(2020)Rathi, Srinivasan, Panda, and Roy}]{rathi2019enabling}
Rathi, N., Srinivasan, G., Panda, P., and Roy, K. (2020).
\newblock Enabling deep spiking neural networks with hybrid conversion and spike timing dependent backpropagation.
\newblock In \emph{International Conference on Learning Representations (ICLR)}
\bibAnnoteFile{rathi2019enabling}

\bibitem[{Ren et~al.(2023{\natexlab{a}})Ren, Ma, Chen, Peng, Liu, Zhang et~al.}]{ren2023spiking}
Ren, D., Ma, Z., Chen, Y., Peng, W., Liu, X., Zhang, Y., et~al. (2023{\natexlab{a}}).
\newblock Spiking pointnet: Spiking neural networks for point clouds.
\newblock In \emph{Advances in Neural Information Processing Systems (NeurIPS)}. vol.~36
\bibAnnoteFile{ren2023spiking}

\bibitem[{Ren et~al.(2023{\natexlab{b}})Ren, Zhou, Huang, Fu, Lin, Song et~al.}]{ren2023spikepoint}
Ren, H., Zhou, Y., Huang, Y., Fu, H., Lin, X., Song, J., et~al. (2023{\natexlab{b}}).
\newblock Spikepoint: An efficient point-based spiking neural network for event cameras action recognition.
\newblock \emph{arXiv preprint arXiv:2310.07189}
\bibAnnoteFile{ren2023spikepoint}

\bibitem[{Roy et~al.(2019)Roy, Jaiswal, and Panda}]{roy2019towards}
Roy, K., Jaiswal, A., and Panda, P. (2019).
\newblock Towards spike-based machine intelligence with neuromorphic computing.
\newblock \emph{Nature} 575, 607--617
\bibAnnoteFile{roy2019towards}

\bibitem[{Rueckauer et~al.(2017)Rueckauer, Lungu, Hu, Pfeiffer, and Liu}]{rueckauer2017conversion}
Rueckauer, B., Lungu, I.-A., Hu, Y., Pfeiffer, M., and Liu, S.-C. (2017).
\newblock Conversion of continuous-valued deep networks to efficient event-driven networks for image classification.
\newblock \emph{Frontiers in Neuroscience} 11, 682
\bibAnnoteFile{rueckauer2017conversion}

\bibitem[{Schemmel et~al.(2010)Schemmel, Br{\"u}derle, Gr{\"u}bl, Hock, Meier, and Millner}]{schemmel2010wafer}
Schemmel, J., Br{\"u}derle, D., Gr{\"u}bl, A., Hock, M., Meier, K., and Millner, S. (2010).
\newblock A wafer-scale neuromorphic hardware system for large-scale neural modeling.
\newblock In \emph{International Symposium on Circuits and Systems (ISCAS)}. 1947--1950
\bibAnnoteFile{schemmel2010wafer}

\bibitem[{Schuman et~al.(2022)Schuman, Plank, and Rose}]{schuman2022application}
Schuman, C., Plank, J., and Rose, G. (2022).
\newblock Application-hardware co-design: System-level optimization of neuromorphic computers with neuromorphic devices.
\newblock In \emph{2022 International Electron Devices Meeting (IEDM)} (IEEE), 2--4
\bibAnnoteFile{schuman2022application}

\bibitem[{Shan et~al.(2022)Shan, Hu, Chen, and Guan}]{shan2022detecting}
Shan, L., Hu, B., Chen, L., and Guan, Z.-H. (2022).
\newblock Detecting covid-19 on ct images with impulsive-backpropagation neural networks.
\newblock In \emph{Chinese Control and Decision Conference (CCDC)}. 2797--2803
\bibAnnoteFile{shan2022detecting}

\bibitem[{Shen et~al.(2023)Shen, Zhao, and Zeng}]{lifb}
Shen, G., Zhao, D., and Zeng, Y. (2023).
\newblock Exploiting high performance spiking neural networks with efficient spiking patterns.
\newblock \emph{arXiv} abs/2301.12356.
\newblock \doi{10.48550/arXiv.2301.12356}
\bibAnnoteFile{lifb}

\bibitem[{Shen et~al.(2016)Shen, Ma, Gu, Zhang, Zhu, Xu et~al.}]{shen2016darwin}
Shen, J., Ma, D., Gu, Z., Zhang, M., Zhu, X., Xu, X., et~al. (2016).
\newblock Darwin: a neuromorphic hardware co-processor based on spiking neural networks.
\newblock \emph{Sci. China Inf. Sci.} 59, 1--5
\bibAnnoteFile{shen2016darwin}

\bibitem[{Shi et~al.(2024)Shi, Hao, and Yu}]{shi2024spikingresformer}
Shi, X., Hao, Z., and Yu, Z. (2024).
\newblock Spikingresformer: Bridging resnet and vision transformer in spiking neural networks.
\newblock In \emph{Proceedings of the IEEE/CVF Conference on Computer Vision and Pattern Recognition (CVPR)}. 5610--5619
\bibAnnoteFile{shi2024spikingresformer}

\bibitem[{Shrestha and Orchard(2018)}]{shrestha2018slayer}
Shrestha, S.~B. and Orchard, G. (2018).
\newblock Slayer: Spike layer error reassignment in time.
\newblock In \emph{Advances in Neural Information Processing Systems (NeurIPS)}. vol.~31
\bibAnnoteFile{shrestha2018slayer}

\bibitem[{Soures and Kudithipudi(2019)}]{soures2019deep}
Soures, N. and Kudithipudi, D. (2019).
\newblock Deep liquid state machines with neural plasticity for video activity recognition.
\newblock \emph{Frontiers in neuroscience} 13, 457929
\bibAnnoteFile{soures2019deep}

\bibitem[{Su et~al.(2023)Su, Chou, Hu, Li, Mei, Zhang et~al.}]{su2023deep}
Su, Q., Chou, Y., Hu, Y., Li, J., Mei, S., Zhang, Z., et~al. (2023).
\newblock Deep directly-trained spiking neural networks for object detection.
\newblock In \emph{Proceedings of the IEEE/CVF International Conference on Computer Vision (ICCV)}. 6555--6565
\bibAnnoteFile{su2023deep}

\bibitem[{Vaswani et~al.(2017)Vaswani, Shazeer, Parmar, Uszkoreit, Jones, Gomez et~al.}]{vaswani2017attention}
Vaswani, A., Shazeer, N., Parmar, N., Uszkoreit, J., Jones, L., Gomez, A.~N., et~al. (2017).
\newblock Attention is all you need.
\newblock In \emph{Advances in Neural Information Processing Systems (NeurIPS)}. vol.~30
\bibAnnoteFile{vaswani2017attention}

\bibitem[{Wang et~al.(2022)Wang, Dong, Zhao, Li, Yang, Yin et~al.}]{Wang2022SpikingED}
Wang, B., Dong, G., Zhao, Y., Li, R., Yang, H., Yin, W., et~al. (2022).
\newblock Spiking emotions: Dynamic vision emotion recognition using spiking neural networks.
\newblock In \emph{Proceedings of the International Conference on Algorithms, High Performance Computing and Artificial Intelligence (AHPCAI)}. vol. 3331, 50--58
\bibAnnoteFile{Wang2022SpikingED}

\bibitem[{Wang and Li(2023)}]{wang2023bioinspiredmem}
Wang, H. and Li, Y.-F. (2023).
\newblock Bioinspired membrane learnable spiking neural network for autonomous vehicle sensors fault diagnosis under open environments.
\newblock \emph{Reliability Engineering \& System Safety} 233, 109102
\bibAnnoteFile{wang2023bioinspiredmem}

\bibitem[{Wang et~al.(2023{\natexlab{a}})Wang, Li, and Zhang}]{wang2023bioinspired}
Wang, H., Li, Y.-F., and Zhang, Y. (2023{\natexlab{a}}).
\newblock Bioinspired spiking spatiotemporal attention framework for lithium-ion batteries state-of-health estimation.
\newblock \emph{Renewable and Sustainable Energy Reviews} 188, 113728
\bibAnnoteFile{wang2023bioinspired}

\bibitem[{Wang et~al.(2023{\natexlab{b}})Wang, Sun, and Li}]{wang2023brain}
Wang, H., Sun, M., and Li, Y.-F. (2023{\natexlab{b}}).
\newblock A brain-inspired spiking network framework based on multi-time-step self-attention for lithium-ion batteries capacity prediction.
\newblock \emph{IEEE Transactions on Consumer Electronics} , 1--1
\bibAnnoteFile{wang2023brain}

\bibitem[{Wang et~al.(2023{\natexlab{c}})Wang, Li, Ma, and Fan}]{wang2023spiking}
Wang, M., Li, J., Ma, M., and Fan, X. (2023{\natexlab{c}}).
\newblock Spiking semantic communication for feature transmission with harq.
\newblock \emph{arXiv} abs/2310.08804.
\newblock \doi{10.48550/arXiv.2310.08804}
\bibAnnoteFile{wang2023spiking}

\bibitem[{Wang and Li(2016)}]{wang2016d}
Wang, Q. and Li, P. (2016).
\newblock D-lsm: Deep liquid state machine with unsupervised recurrent reservoir tuning.
\newblock In \emph{2016 23rd International Conference on Pattern Recognition (ICPR)} (IEEE), 2652--2657
\bibAnnoteFile{wang2016d}

\bibitem[{Wang et~al.(2017)Wang, Thakur, Cohen, Hamilton, Tapson, and van Schaik}]{wang2017neuromorphic}
Wang, R., Thakur, C.~S., Cohen, G., Hamilton, T.~J., Tapson, J., and van Schaik, A. (2017).
\newblock Neuromorphic hardware architecture using the neural engineering framework for pattern recognition.
\newblock \emph{IEEE Transactions on Biomedical Circuits and Systems} 11, 574--584
\bibAnnoteFile{wang2017neuromorphic}

\bibitem[{WANG et~al.(2022)WANG, Cheng, and Lim}]{ltmd}
WANG, S., Cheng, T.~H., and Lim, M.-H. (2022).
\newblock Ltmd: Learning improvement of spiking neural networks with learnable thresholding neurons and moderate dropout.
\newblock In \emph{Advances in Neural Information Processing Systems (NeurIPS)}. vol.~35, 28350--28362
\bibAnnoteFile{ltmd}

\bibitem[{Wang et~al.(2019)Wang, Hao, Wei, Xiao, Feng, and Sebe}]{wang2019temporal}
Wang, W., Hao, S., Wei, Y., Xiao, S., Feng, J., and Sebe, N. (2019).
\newblock Temporal spiking recurrent neural network for action recognition.
\newblock \emph{IEEE Access} 7, 117165--117175
\bibAnnoteFile{wang2019temporal}

\bibitem[{Wang et~al.(2021)Wang, Xie, Li, Fan, Song, Liang et~al.}]{wang2021pyramid}
Wang, W., Xie, E., Li, X., Fan, D.-P., Song, K., Liang, D., et~al. (2021).
\newblock Pyramid vision transformer: A versatile backbone for dense prediction without convolutions.
\newblock In \emph{Proceedings of the IEEE/CVF International Conference on Computer Vision (ICCV)}. 568--578
\bibAnnoteFile{wang2021pyramid}

\bibitem[{Wang et~al.(2023{\natexlab{d}})Wang, Zhang, and Zhang}]{mt-snn}
Wang, X., Zhang, Y., and Zhang, Y. (2023{\natexlab{d}}).
\newblock Mt-snn: Enhance spiking neural network with multiple thresholds.
\newblock \emph{arXiv} abs/2303.11127.
\newblock \doi{10.48550/arXiv.2303.11127}
\bibAnnoteFile{mt-snn}

\bibitem[{Wang et~al.(2023{\natexlab{e}})Wang, Shi, Lu, Liu, Zhang, and Qu}]{ijcai2023p344}
Wang, Y., Shi, K., Lu, C., Liu, Y., Zhang, M., and Qu, H. (2023{\natexlab{e}}).
\newblock Spatial-temporal self-attention for asynchronous spiking neural networks.
\newblock In \emph{Proceedings of the Thirty-Second International Joint Conference on Artificial Intelligence (IJCAI)}. 3085--3093
\bibAnnoteFile{ijcai2023p344}

\bibitem[{Wang et~al.(2022)Wang, Zhang, Chen, and Qu}]{wang2022signed}
Wang, Y., Zhang, M., Chen, Y., and Qu, H. (2022).
\newblock Signed neuron with memory: Towards simple, accurate and high-efficient ann-snn conversion.
\newblock In \emph{Proceedings of the Thirty-First International Joint Conference on Artificial Intelligence (IJCAI))}. 2501--2508
\bibAnnoteFile{wang2022signed}

\bibitem[{Werbos(1990)}]{BPTT_original}
Werbos, P. (1990).
\newblock Backpropagation through time: what it does and how to do it.
\newblock \emph{Proceedings of the IEEE} 78, 1550--1560.
\newblock \doi{10.1109/5.58337}
\bibAnnoteFile{BPTT_original}

\bibitem[{Wiener and Richmond(2003)}]{Wiener2394}
Wiener, M.~C. and Richmond, B.~J. (2003).
\newblock Decoding spike trains instant by instant using order statistics and the mixture-of-poissons model.
\newblock \emph{Journal of Neuroscience} 23, 2394--2406
\bibAnnoteFile{Wiener2394}

\bibitem[{Wu et~al.(2020)Wu, Huang, Zhu, Liang, Hong, Deng et~al.}]{wu2020compact}
Wu, M.-H., Huang, M.-S., Zhu, Z., Liang, F.-X., Hong, M.-C., Deng, J., et~al. (2020).
\newblock Compact probabilistic poisson neuron based on back-hopping oscillation in stt-mram for all-spin deep spiking neural network.
\newblock In \emph{Symposium on VLSI Technology}. 1--2
\bibAnnoteFile{wu2020compact}

\bibitem[{Wu et~al.(2022)Wu, He, Yao, Zhang, Wang, and Li}]{wu2022mss}
Wu, X., He, W., Yao, M., Zhang, Z., Wang, Y., and Li, G. (2022).
\newblock Mss-depthnet: Depth prediction with multi-step spiking neural network.
\newblock \emph{arXiv} abs/2211.12156.
\newblock \doi{10.48550/arXiv.2211.12156}
\bibAnnoteFile{wu2022mss}

\bibitem[{Wu et~al.(2018)Wu, Deng, Li, Zhu, and Shi}]{wu2018spatio}
Wu, Y., Deng, L., Li, G., Zhu, J., and Shi, L. (2018).
\newblock Spatio-temporal backpropagation for training high-performance spiking neural networks.
\newblock \emph{Frontiers in Neuroscience} 12, 331
\bibAnnoteFile{wu2018spatio}

\bibitem[{Wu et~al.(2019)Wu, Deng, Li, Zhu, Xie, and Shi}]{wu2019direct}
Wu, Y., Deng, L., Li, G., Zhu, J., Xie, Y., and Shi, L. (2019).
\newblock Direct training for spiking neural networks: Faster, larger, better.
\newblock In \emph{Proceedings of the AAAI Conference on Artificial Intelligence (AAAI)}. 1311--1318
\bibAnnoteFile{wu2019direct}

\bibitem[{Xiang et~al.(2022)Xiang, Zhang, Jiang, Han, Zhang, Du et~al.}]{xiang2022spiking}
Xiang, S., Zhang, T., Jiang, S., Han, Y., Zhang, Y., Du, C., et~al. (2022).
\newblock Spiking siamfc++: Deep spiking neural network for object tracking.
\newblock \emph{arXiv} abs/2209.12010.
\newblock \doi{10.48550/arXiv.2209.12010}
\bibAnnoteFile{xiang2022spiking}

\bibitem[{Xiao et~al.(2022)Xiao, Meng, Zhang, He, and Lin}]{OTTT}
Xiao, M., Meng, Q., Zhang, Z., He, D., and Lin, Z. (2022).
\newblock Online training through time for spiking neural networks.
\newblock In \emph{Advances in Neural Information Processing Systems (NeurIPS)}. vol.~35, 20717--20730
\bibAnnoteFile{OTTT}

\bibitem[{Xiong et~al.(2021)Xiong, Chen, Chen, Wei, Xue, Wan et~al.}]{xiong2021odor}
Xiong, Y., Chen, Y., Chen, C., Wei, X., Xue, Y., Wan, H., et~al. (2021).
\newblock An odor recognition algorithm of electronic noses based on convolutional spiking neural network for spoiled food identification.
\newblock \emph{Journal of the Electrochemical Society} 168, 077519
\bibAnnoteFile{xiong2021odor}

\bibitem[{Xu et~al.(2022)Xu, Ma, Pan, and Zheng}]{xu2022deep}
Xu, Z., Ma, Y., Pan, Z., and Zheng, X. (2022).
\newblock Deep spiking residual shrinkage network for bearing fault diagnosis.
\newblock \emph{IEEE Transactions on Cybernetics} , 1--6
\bibAnnoteFile{xu2022deep}

\bibitem[{Yamazaki et~al.(2022)Yamazaki, Vo-Ho, Bulsara, and Le}]{yamazaki2022spiking}
Yamazaki, K., Vo-Ho, V.-K., Bulsara, D., and Le, N. (2022).
\newblock Spiking neural networks and their applications: A review.
\newblock \emph{Brain Sciences} 12, 863
\bibAnnoteFile{yamazaki2022spiking}

\bibitem[{Yang et~al.(2019)Yang, Wu, Wang, Yang, Li, Deng et~al.}]{yang2019dashnet}
Yang, Z., Wu, Y., Wang, G., Yang, Y., Li, G., Deng, L., et~al. (2019).
\newblock Dashnet: {A} hybrid artificial and spiking neural network for high-speed object tracking.
\newblock \emph{arXiv} abs/1909.12942.
\newblock \doi{10.48550/arXiv.1909.12942}
\bibAnnoteFile{yang2019dashnet}

\bibitem[{Yao et~al.(2024)Yao, Hu, Hu, Xu, Zhou, Tian et~al.}]{yao2024spikedriven}
Yao, M., Hu, J., Hu, T., Xu, Y., Zhou, Z., Tian, Y., et~al. (2024).
\newblock Spike-driven transformer v2: Meta spiking neural network architecture inspiring the design of next-generation neuromorphic chips.
\newblock In \emph{The Twelfth International Conference on Learning Representations (ICLR)}
\bibAnnoteFile{yao2024spikedriven}

\bibitem[{Yao et~al.(2023{\natexlab{a}})Yao, Hu, Zhou, Yuan, Tian, Xu et~al.}]{yao2023spikedriven}
Yao, M., Hu, J., Zhou, Z., Yuan, L., Tian, Y., Xu, B., et~al. (2023{\natexlab{a}}).
\newblock Spike-driven transformer.
\newblock In \emph{Advances in Neural Information Processing Systems (NeurIPS)}. vol.~36
\bibAnnoteFile{yao2023spikedriven}

\bibitem[{Yao et~al.(2023{\natexlab{b}})Yao, Zhao, Zhang, Hu, Deng, Tian et~al.}]{10032591}
Yao, M., Zhao, G., Zhang, H., Hu, Y., Deng, L., Tian, Y., et~al. (2023{\natexlab{b}}).
\newblock Attention spiking neural networks.
\newblock \emph{IEEE Transactions on Pattern Analysis and Machine Intelligence} 45, 9393--9410
\bibAnnoteFile{10032591}

\bibitem[{Yao et~al.(2022)Yao, Li, Mo, and Cheng}]{glif}
Yao, X., Li, F., Mo, Z., and Cheng, J. (2022).
\newblock Glif: A unified gated leaky integrate-and-fire neuron for spiking neural networks.
\newblock In \emph{Advances in Neural Information Processing Systems (NeurIPS)}. vol.~35, 32160--32171
\bibAnnoteFile{glif}

\bibitem[{Yarga and Wood(2023)}]{spsn}
Yarga, S. Y.~A. and Wood, S. U.~N. (2023).
\newblock Accelerating snn training with stochastic parallelizable spiking neurons.
\newblock In \emph{International Joint Conference on Neural Networks (IJCNN)}. 1--8
\bibAnnoteFile{spsn}

\bibitem[{Yin et~al.(2020)Yin, adi, and Boht'e}]{Yin2020EffectiveAE}
Yin, B., adi, F., and Boht'e, S.~M. (2020).
\newblock Effective and efficient computation with multiple-timescale spiking recurrent neural networks.
\newblock \emph{International Conference on Neuromorphic Systems 2020} , 1–8
\bibAnnoteFile{Yin2020EffectiveAE}

\bibitem[{Yin et~al.(2022)Yin, Moitra, Bhattacharjee, Kim, and Panda}]{yin2022sata}
Yin, R., Moitra, A., Bhattacharjee, A., Kim, Y., and Panda, P. (2022).
\newblock Sata: Sparsity-aware training accelerator for spiking neural networks.
\newblock \emph{IEEE Transactions on Computer-Aided Design of Integrated Circuits and Systems} 42, 1926 -- 1938
\bibAnnoteFile{yin2022sata}

\bibitem[{Yuan et~al.(2021)Yuan, Chen, Wang, Yu, Shi, Jiang et~al.}]{yuan2021tokens}
Yuan, L., Chen, Y., Wang, T., Yu, W., Shi, Y., Jiang, Z.-H., et~al. (2021).
\newblock Tokens-to-token vit: Training vision transformers from scratch on imagenet.
\newblock In \emph{Proceedings of the IEEE/CVF International Conference on Computer Vision (ICCV)}. 558--567
\bibAnnoteFile{yuan2021tokens}

\bibitem[{Yuan et~al.(2022)Yuan, Hou, Jiang, Feng, and Yan}]{yuan2022volo}
Yuan, L., Hou, Q., Jiang, Z., Feng, J., and Yan, S. (2022).
\newblock Volo: Vision outlooker for visual recognition.
\newblock \emph{IEEE Transactions on Pattern Analysis and Machine Intelligence} 45, 6575--6586
\bibAnnoteFile{yuan2022volo}

\bibitem[{Zhang et~al.(2023{\natexlab{a}})Zhang, Fan, and Zhang}]{zhang2023energy}
Zhang, H., Fan, X., and Zhang, Y. (2023{\natexlab{a}}).
\newblock Energy-efficient spiking segmenter for frame and event-based images.
\newblock \emph{Biomimetics} 8, 356
\bibAnnoteFile{zhang2023energy}

\bibitem[{Zhang et~al.(2024)Zhang, Zhou, Yu, Huang, Ma, Fan et~al.}]{zhang2024sglformer}
Zhang, H., Zhou, C., Yu, L., Huang, L., Ma, Z., Fan, X., et~al. (2024).
\newblock Sglformer: Spiking global-local-fusion transformer with high performance.
\newblock \emph{Frontiers in Neuroscience} 18, 1371290
\bibAnnoteFile{zhang2024sglformer}

\bibitem[{Zhang et~al.(2022{\natexlab{a}})Zhang, Dong, Zhang, Ding, Heide, Yin et~al.}]{zhang2022spiking}
Zhang, J., Dong, B., Zhang, H., Ding, J., Heide, F., Yin, B., et~al. (2022{\natexlab{a}}).
\newblock Spiking transformers for event-based single object tracking.
\newblock In \emph{Proceedings of the IEEE/CVF Conference on Computer Vision and Pattern Recognition (CVPR)}. 8801--8810
\bibAnnoteFile{zhang2022spiking}

\bibitem[{Zhang et~al.(2023{\natexlab{b}})Zhang, Huang, Ma, and Zhou}]{zhang2023predicting}
Zhang, J., Huang, L., Ma, Z., and Zhou, H. (2023{\natexlab{b}}).
\newblock Predicting the temporal-dynamic trajectories of cortical neuronal responses in non-human primates based on deep spiking neural network.
\newblock \emph{Cognitive Neurodynamics} , 1--12
\bibAnnoteFile{zhang2023predicting}

\bibitem[{Zhang et~al.(2022{\natexlab{b}})Zhang, Tang, Yu, Lu, and Huang}]{zhang2022spike}
Zhang, J., Tang, L., Yu, Z., Lu, J., and Huang, T. (2022{\natexlab{b}}).
\newblock Spike transformer: Monocular depth estimation for spiking camera.
\newblock In \emph{Proceedings of the European Conference on Computer Vision (ECCV)}. vol. 13667, 34--52
\bibAnnoteFile{zhang2022spike}

\bibitem[{Zhang et~al.(2022{\natexlab{c}})Zhang, Wang, Di, and Pu}]{zhang2022high}
Zhang, J., Wang, J., Di, X., and Pu, S. (2022{\natexlab{c}}).
\newblock High-accuracy and energy-efficient action recognition with deep spiking neural network.
\newblock In \emph{International Conference on Neural Information Processing (ICONIP)}. 279--292
\bibAnnoteFile{zhang2022high}

\bibitem[{Zhang et~al.(2021)Zhang, Lu, Wang, Wang, Wei, Shi et~al.}]{zhang2021hybrid}
Zhang, X., Lu, J., Wang, Z., Wang, R., Wei, J., Shi, T., et~al. (2021).
\newblock Hybrid memristor-cmos neurons for in-situ learning in fully hardware memristive spiking neural networks.
\newblock \emph{Science Bulletin} 66, 1624--1633
\bibAnnoteFile{zhang2021hybrid}

\bibitem[{Zhang et~al.(2022{\natexlab{d}})Zhang, Zhang, and Lew}]{zhang2022pokebnn}
Zhang, Y., Zhang, Z., and Lew, L. (2022{\natexlab{d}}).
\newblock Pokebnn: A binary pursuit of lightweight accuracy.
\newblock In \emph{Proceedings of the IEEE/CVF Conference on Computer Vision and Pattern Recognition (CVPR)}. 12475--12485
\bibAnnoteFile{zhang2022pokebnn}

\bibitem[{Zheng et~al.(2021)Zheng, Wu, Deng, Hu, and Li}]{zheng2021going}
Zheng, H., Wu, Y., Deng, L., Hu, Y., and Li, G. (2021).
\newblock Going deeper with directly-trained larger spiking neural networks.
\newblock In \emph{Proceedings of the AAAI Conference on Artificial Intelligence (AAAI)}. 11062--11070
\bibAnnoteFile{zheng2021going}

\bibitem[{Zhou et~al.(2023{\natexlab{a}})Zhou, Yu, Zhou, Zhang, Ma, Zhou et~al.}]{zhou2023spikingformer}
Zhou, C., Yu, L., Zhou, Z., Zhang, H., Ma, Z., Zhou, H., et~al. (2023{\natexlab{a}}).
\newblock Spikingformer: Spike-driven residual learning for transformer-based spiking neural network.
\newblock \emph{arXiv} abs/2304.11954.
\newblock \doi{10.48550/arXiv.2304.11954}
\bibAnnoteFile{zhou2023spikingformer}

\bibitem[{Zhou et~al.(2024{\natexlab{a}})Zhou, Zhang, Zhou, Yu, Huang, Fan et~al.}]{zhou2024qkformer}
Zhou, C., Zhang, H., Zhou, Z., Yu, L., Huang, L., Fan, X., et~al. (2024{\natexlab{a}}).
\newblock Qkformer: Hierarchical spiking transformer using qk attention.
\newblock \emph{arXiv preprint arXiv:2403.16552}
\bibAnnoteFile{zhou2024qkformer}

\bibitem[{Zhou et~al.(2023{\natexlab{b}})Zhou, Zhang, Zhou, Yu, Ma, Zhou et~al.}]{zhou2023enhancing}
Zhou, C., Zhang, H., Zhou, Z., Yu, L., Ma, Z., Zhou, H., et~al. (2023{\natexlab{b}}).
\newblock Enhancing the performance of transformer-based spiking neural networks by snn-optimized downsampling with precise gradient backpropagation.
\newblock \emph{arXiv} abs/2305.05954.
\newblock \doi{10.48550/arXiv.2305.05954}
\bibAnnoteFile{zhou2023enhancing}

\bibitem[{Zhou et~al.(2020)Zhou, Chen, Li, and Sanyal}]{zhou2020deep}
Zhou, S., Chen, Y., Li, X., and Sanyal, A. (2020).
\newblock Deep scnn-based real-time object detection for self-driving vehicles using lidar temporal data.
\newblock \emph{IEEE Access} 8, 76903--76912
\bibAnnoteFile{zhou2020deep}

\bibitem[{Zhou et~al.(2024{\natexlab{b}})Zhou, Che, Fang, Tian, Zhu, Yan et~al.}]{zhou2024spikformerv2}
Zhou, Z., Che, K., Fang, W., Tian, K., Zhu, Y., Yan, S., et~al. (2024{\natexlab{b}}).
\newblock Spikformer v2: Join the high accuracy club on imagenet with an snn ticket.
\newblock \emph{arXiv} abs/2401.02020.
\newblock \doi{10.48550/arXiv.2401.02020}
\bibAnnoteFile{zhou2024spikformerv2}

\bibitem[{Zhou et~al.(2023{\natexlab{c}})Zhou, Zhu, He, Wang, YAN, Tian et~al.}]{zhou2023spikformer}
Zhou, Z., Zhu, Y., He, C., Wang, Y., YAN, S., Tian, Y., et~al. (2023{\natexlab{c}}).
\newblock Spikformer: When spiking neural network meets transformer.
\newblock In \emph{International Conference on Learning Representations (ICLR)}
\bibAnnoteFile{zhou2023spikformer}

\bibitem[{Zhu et~al.(2021{\natexlab{a}})Zhu, Li, Wang, Huang, and Tian}]{zhu2021neuspike}
Zhu, L., Li, J., Wang, X., Huang, T., and Tian, Y. (2021{\natexlab{a}}).
\newblock Neuspike-net: High speed video reconstruction via bio-inspired neuromorphic cameras.
\newblock In \emph{Proceedings of the IEEE/CVF International Conference on Computer Vision (ICCV)}. 2400--2409
\bibAnnoteFile{zhu2021neuspike}

\bibitem[{Zhu and Tian(2023)}]{zhu2023revrec}
Zhu, L. and Tian, Y. (2023).
\newblock Review of visual reconstruction methods of retina-like vision sensors.
\newblock \emph{Scientia Sinica (Informationis)} 53, 417--436
\bibAnnoteFile{zhu2023revrec}

\bibitem[{Zhu et~al.(2022{\natexlab{a}})Zhu, Wang, Chang, Li, Huang, and Tian}]{zhu2022event}
Zhu, L., Wang, X., Chang, Y., Li, J., Huang, T., and Tian, Y. (2022{\natexlab{a}}).
\newblock Event-based video reconstruction via potential-assisted spiking neural network.
\newblock In \emph{Proceedings of the IEEE/CVF Conference on Computer Vision and Pattern Recognition (CVPR)}. 3594--3604
\bibAnnoteFile{zhu2022event}

\bibitem[{Zhu et~al.(2023)Zhu, Zhao, and Eshraghian}]{zhu2023spikegpt}
Zhu, R., Zhao, Q., and Eshraghian, J.~K. (2023).
\newblock Spikegpt: Generative pre-trained language model with spiking neural networks.
\newblock \emph{arXiv} abs/2302.13939.
\newblock \doi{10.48550/arXiv.2302.13939}
\bibAnnoteFile{zhu2023spikegpt}

\bibitem[{Zhu et~al.(2021{\natexlab{b}})Zhu, Su, Lu, Li, Wang, and Dai}]{zhu2020deformable}
Zhu, X., Su, W., Lu, L., Li, B., Wang, X., and Dai, J. (2021{\natexlab{b}}).
\newblock Deformable detr: Deformable transformers for end-to-end object detection.
\newblock In \emph{International Conference on Learning Representations (ICLR)}
\bibAnnoteFile{zhu2020deformable}

\bibitem[{Zhu et~al.(2022{\natexlab{b}})Zhu, Peng, Li, Chen, Yu, and Luo}]{zhu2022spiking}
Zhu, Z., Peng, J., Li, J., Chen, L., Yu, Q., and Luo, S. (2022{\natexlab{b}}).
\newblock Spiking graph convolutional networks.
\newblock \emph{arXiv preprint arXiv:2205.02767}
\bibAnnoteFile{zhu2022spiking}

\bibitem[{Zou et~al.(2023)Zou, Mu, Zuo, Wang, and Li}]{zou2023event}
Zou, S., Mu, Y., Zuo, X., Wang, S., and Li, C. (2023).
\newblock Event-based human pose tracking by spiking spatiotemporal transformer.
\newblock \emph{arXiv} abs/2303.09681.
\newblock \doi{10.48550/arXiv.2303.09681}
\bibAnnoteFile{zou2023event}

\end{thebibliography}

\end{document}